\def\doctitle{A Cooperative Coordination Solver for Travelling Thief Problems}
\def\docauthor{Majid Namazi, Conrad Sanderson, M.A. Hakim Newton, Abdul Sattar}

\documentclass[nopreprintline,review,a4paper]{elsarticle}

\usepackage[pagebackref=false,breaklinks=true,colorlinks,bookmarks=false,urlcolor=blue,linkcolor=blue,citecolor=blue,pdftitle={\doctitle},pdfauthor={\docauthor}]{hyperref}
\usepackage[includefoot,includehead,a4paper,top=0.5cm,bottom=1cm,left=2cm,right=2cm]{geometry}

\usepackage{times}
\usepackage{helvet}
\usepackage{courier}
\usepackage{subcaption}
\usepackage{graphicx}
\usepackage{amsmath}
\usepackage{amssymb}
\usepackage{algorithm}
\usepackage{algorithmic}
\usepackage{multirow}
\usepackage{adjustbox}
\usepackage{color}
\usepackage{pdflscape}
\usepackage[british]{babel}
\small\normalsize

\begin{document}

\pagestyle{empty}

\begin{frontmatter}

\title{A Cooperative Coordination Solver for Travelling Thief Problems}

\author
  {
  Majid Namazi~{$^{1,2}$},
  Conrad Sanderson~$^{1,2}$,
  M.A. Hakim Newton~$^{2}$,
  Abdul Sattar~$^{2}$
  }

\address
  {
  $^{1}$ Data61~/~CSIRO, Australia\\
  $^{2}$ Griffith University, Australia
  }

\begin{abstract}

The travelling thief problem (TTP) is a representative of multi-component optimisation problems
where the components interact with each other.
TTP combines two interdependent components:
the knapsack problem (KP) and the travelling salesman problem (TSP).
A thief performs a cyclic tour through a set of cities,
and pursuant to a collection plan,
collects a subset of items into a rented knapsack with finite capacity.
The aim is to maximise overall profit while minimising renting cost.
Existing TTP solvers typically solve the KP and TSP components in an interleaved manner,
where the solution of one component is kept fixed while the solution of the other component is modified.
This suggests low coordination between solving the two components,
possibly leading to low-quality TTP solutions.
The \mbox{2-OPT} heuristic is often used for solving the TSP component,
which reverses a segment in the cyclic tour.
Within the TTP context, \mbox{2-OPT} does not take into account the collection plan,
which can result in a lower objective value.
This in turn can result in the tour modification to be rejected by a solver.
To address this issue,
we propose an expanded form of 2-OPT in order to change the collection plan
in coordination with modifying the tour.
Items regarded as less profitable and collected in cities located earlier in the reversed segment
are substituted by items that tend to be more profitable and not collected in cities
located later in the reversed segment.
The collection plan is further changed through a modified form of the hill-climbing bit-flip search,
where changes in the collection state are only permitted for boundary items,
which are defined as lowest profitable collected items or highest profitable uncollected items.
This restriction reduces the amount of time spent on the KP component,
thereby allowing more tours to be evaluated by the TSP component within a given time budget.
The two proposed approaches form the basis of a new cooperative coordination solver,
which finds a large number of new best gain values
and outperforms several state-of-the-art TTP solvers on a comprehensive set of benchmark TTP instances.%
\footnote{This article is a revised and extended version of our earlier work~\cite{namazi2019pgch}.}

\vspace{1.5ex}


\end{abstract}

\end{frontmatter}

\section{Introduction}

Practical constraint optimisation problems~\cite{rossi2006handbook}
often consist of several interdependent components~\cite{bonyadi2013travelling}.
Finding an optimal overall solution to such problems
cannot be guaranteed by simply finding an optimal solution to each underlying component~\cite{michalewicz2012quo,mei2016investigation,bonyadi2019evolutionary}.
A representative of such multi-component problems is the travelling thief problem (TTP)~\cite{bonyadi2013travelling,polyakovskiy2014comprehensive},
which comprises two interdependent NP-hard components:
the knapsack problem (KP)~\cite{inbookKP}
and the travelling salesman problem (TSP)~\cite{gutin2006traveling}.
The overall problem is described as follows.
A thief performs a {\em cyclic tour} through a set of specified cities.
The thief collects a subset of obtainable items into a finite capacity knapsack 
pursuant to a {\em collection plan}.
The knapsack is rented, so it has an associated cost.
As items are collected at each city,
the overall weight and profit of the items in the knapsack increase,
leading to a decrease in the speed of the thief.
This in turn leads to an increase to the overall travelling time and thus the cost of the knapsack.
The aim of a TTP solution is to concurrently minimise the renting cost and maximise the overall profit of the collected items.
TTP can be thought of as a proxy for many real-world logistics problems~\cite{mei2014improving}.

Existing TTP solvers typically use separate modules for solving the KP and TSP components,
and follow an interleaved approach:
the solution of one component is kept fixed while the solution of the other component is modified~\cite{polyakovskiy2014comprehensive}.
In this context, 
the aim of solving the KP component is to maximise the overall profit given a fixed cyclic tour,
while 
the aim of solving the TSP component is to minimise the overall travelling time given a fixed collection plan.
However, this interleaved approach
suggests low coordination between solving the two components, possibly leading to low quality TTP solutions.

The 2-OPT segment reversal heuristic~\cite{croes1958method} is often employed for solving the TSP component.
In the context of TTP, the 2-OPT heuristic does not take into account the collection plan,
which can result in a decrease of the TTP objective value.
This in turn can result in the tour modification to be rejected by a solver,
which suggests that many possible segment reversals can be rejected without considering potential changes to the collection plan.
This unnecessarily restricts the search of the TTP solution space.

For solving the KP component, a popular approach is the bit-flip search~\cite{polyakovskiy2014comprehensive,faulkner2015approximate},
which is a hill-climber that searches via flipping the collection status (from collected to uncollected and vice-versa) of one item at a time.
The downside of this approach is that the small and untargeted change in the collection plan results in a slow and meandering exploration of the solution space.
When a time limit is placed for finding a solution, the solution space may not be explored adequately,
which can contribute to low quality TTP solutions.

To address the poor coordination between solving the KP and TSP components,
we propose a modified and extended version of 2-OPT,
in order to explicitly adjust the collection plan in coordination with changes made to the cyclic tour.
After reversing the segment, items regarded as less profitable and collected in cities located earlier in the reversed segment
are substituted by items that tend to be equally or more profitable and not collected in cities located later in the reversed segment.
We term the new heuristic as Profit Guided Coordination Heuristic (PGCH).
An early version of the PGCH heuristic, proposed in our preliminary work~\cite{namazi2019pgch},
uses only the sequence of (the profitability ratios of) the lowest profitable collected items to adjust the collection plan.
In contrast, the extended version uses both the sequence of the lowest profitable collected items and the sequence of the highest profitable uncollected items
to adjust the collection plan, as discussed in more detail in Section~\ref{sec:pgch}. 

We further extend our preliminary work~\cite{namazi2019pgch} by proposing a more targeted form of the bit-flip search,
termed Boundary Bit-Flip search,
where a restriction is placed to only consider changes to the collection state of {\it boundary items}.
We define two types of boundary items:
(i) lowest profitable collected items among all items collected earlier in the segment,
and
(ii) highest profitable uncollected items among all items not collected later in the segment.
The restriction reduces the amount of time spent on the KP component,
thereby allowing more tours to be evaluated by the TSP component within a time limit.
We combine the proposed PGCH and Boundary Bit-Flip approaches into a new TTP solver,
termed as cooperative coordination (CoCo) solver.
A comparative evaluation on a comprehensive set of benchmark TTP instances
shows that the proposed solver finds a large number of new best gain values and also outperforms several state-of-the-art TTP solvers:
MATLS~\cite{mei2014improving}, S5~\cite{faulkner2015approximate}, CS2SA*~\cite{el2018efficiently} and MEA2P~\cite{wuijts2019investigation}.

We continue the paper as follows.
Section~\ref{sec:related} provides an overview of related work.
Section~\ref{sec:background} formally defines TTP, the 2-OPT heuristic, the bit-flip operator, and other necessary mathematical details.
Section~\ref{sec:pgch} describes the proposed coordination heuristic.
Section~\ref{sec:boundary} describes the proposed targeted form of bit-flip search.
Section~\ref{sec:solver} combines the two proposed heuristics into the proposed CoCo solver.
Section~\ref{sec:experiments} empirically shows the effects of the proposed heuristics and provides the comparative evaluation.
Section~\ref{sec:conclusion} summarises the main findings.

\section{Related Work}
\label{sec:related}

TTP was introduced in~\cite{bonyadi2013travelling}
with many benchmark instances given in~\cite{polyakovskiy2014comprehensive}.
Existing TTP solvers can be grouped into 5~main categories:
{(i)}~constructive methods, 
{(ii)}~fixed-tour methods,
{(iii)}~cooperative methods,
{(iv)}~full encoding methods,
and
{(v)}~hyper-heuristic methods.
Each of the categories is briefly overviewed below.
For a more thorough treatment, the reader is directed to the recent review of TTP solvers in~\cite{wagner2017case}.

In constructive methods, 
an initial cyclic tour is generated for the TSP component using the classic
Chained Lin-Kernighan heuristic~\cite{applegate2003chained}.
The tour is then kept fixed while the collection plan for the KP component is generated
by employing scores assigned to the items based on their weight, profit, and position in the tour.
This category includes approaches such as
Simple Heuristic~\cite{polyakovskiy2014comprehensive},
Density-based Heuristic~\cite{bonyadi2014socially},
Insertion~\cite{mei2014improving}
and 
PackIterative~\cite{faulkner2015approximate}.
These approaches are used in restart-based algorithms such as S5~\cite{faulkner2015approximate}
and in the initialisation phase of more elaborate methods.

In fixed-tour methods,
after generating an initial cyclic tour as per constructive methods,
an iterative improvement heuristic is used to solve the KP component. 
Two iterative methods for solving the KP component are proposed in~\cite{polyakovskiy2014comprehensive}:
(i) Random Local Search,
which is a hill-climbing bit-flip search where the collection status of a randomly selected item is flipped in each iteration,
and
(ii) \mbox{(1+1)-EA},
a simple evolutionary algorithm where the collection status of a set of randomly selected items is flipped in each iteration. 

Cooperative methods 
are iterative approaches based on co-operational co-evolution~\cite{potter1994cooperative}.
After generating an initial TTP solution using a constructive or fixed-tour method, 
the KP and TSP components are solved by two separate modules.
These two modules are executed by a coordinating agent (or meta-optimiser) 
in an interleaved form.
The coordinating agent combines the two solutions to produce an overall solution,
thereby considering the interdependency between the KP and TSP components~\cite{wagner2017case}.
Example methods include
CoSolver~\cite{bonyadi2014socially},
CoSolver with 2-OPT and Simulated Annealing (CS2SA)~\cite{el2016population},
and CS2SA with offline instance-based parameter tuning (CS2SA*)~\cite{el2018efficiently}.

In full-encoding methods, 
the problem is considered as a whole.
Example methods include
Memetic Algorithm with Two-stage Local Search (MATLS)~\cite{mei2014improving},
memetic algorithm with edge-assembly and two-point crossover operators (MEA2P)~\cite{wuijts2019investigation}, 
a swarm intelligence algorithm~\cite{wagner2016stealing} based on max\textendash min ant system~\cite{stutzle2000max},
Memetic Algorithm with 2-OPT and Bit-Flip search~\cite{el2016population},
and 
Joint 2-OPT and Bit-Flip~\cite{el2017local},
which changes the collection status of just one item each time a segment in the cyclic tour is reversed.

In hyper-heuristic based methods,
genetic programming is used to generate or select low level heuristics for the TSP and/or KP components.
In~\cite{mei2015heuristic}, a genetic programming based approach generates two packing heuristics for the KP component.
An individual in each generation is a tree whose internal nodes are simple arithmetic operators,
while the leaf nodes are the numerical parameters of a given TTP instance.
In~\cite{martins2017hseda,el2018hyperheuristic},
genetic programming is used to learn how to select a sequence of low level heuristics
to address both the KP and TSP components.
In~\cite{martins2017hseda},
an individual in each generation is a Bayesian network in which each node corresponds to a low level heuristic.
In~\cite{el2018hyperheuristic},
an individual in each generation is a tree in which the internal nodes are 
functions while the leaf nodes correspond to low level heuristics.

\section{Background}
\label{sec:background}

Each TTP instance is comprised of
a set {$\{1,\ldots,m\}$} of {$m$} items 
and 
a set {$\{1,\ldots,n\}$} of {$n$} cities.
The {\em distance} between each pair of cities {$i\ne i'$} is {$d(i,i') = d(i',i)$}.
Each {\em item} {$j$} is located at {\em city} {$l_j > 1$}
(ie.,~there are no items in the city {$1$}).
Moreover, each item has {\em weight} {$w_j> 0$},
{\em profit} {$\pi_j > 0$}, and corresponding {\em profitability ratio} {$r_j = \pi_j/w_j$}.
An item {$j$} is regarded {\em more profitable} than item {$j'$}
if {$r_j > r_{j'}$}, or {$\pi_j > \pi_{j'}$} if {$r_j = r_{j'}$}.

The thief begins a {\em cyclic tour} at city {$1$},
travels between cities (visiting each city only once),
collects a subset of the items available in each city,
and returns to city {$1$} at the conclusion of the tour.
The cyclic tour is represented by using a permutation of {$n$} cities.
Let us represent a given cyclic tour as {$c$},
with {$c_k = i$} indicating that the {$k$}-th city in the tour {$c$} is {$i$},
and {$c(i) = k$} indicating that the position of city {$i$} in the tour {$c$} is {$k$}.
Here {$c_1=1$} and {$c(1)=1$}. 
A knapsack with a rent rate {$R$} per unit time and a weight capacity {$W$}
is rented by the thief to hold the collected items.
A {\em collection plan} {$p$} indicates that item $j$ is selected for collection if {$p_j=1$},
or not selected if {$p_j=0$}.
An overall solution that provides a cyclic tour {$c$} and a collection plan {$p$} is expressed as {$\langle c,p \rangle$}.

The combined weight of the items collected from city {$i$} is denoted by {\small $W_p(i) = \sum_{l_j= i}w_jp_j$}.
The combined weight of the items collected from the initial {$k$} cities in the tour {$c$}
is denoted by {\small $W_{c,p}(k) = \sum_{k'=1}^{k}W_p(c_{k'})$}.
The thief traverses from city {$c_k$} to the next city with speed {$v_{c,p}(k)$}.
The speed decreases as {\small $W_{c,p}(k)$} increases.
The speed at the city {$c_k$} is given by {\small $v_{c,p}(k) = v_\textrm{max} - W_{c,p}(k)\times(v_\textrm{max} - v_\textrm{min})/W$},
where {$v_\textrm{min}$} and {$v_\textrm{max}$} are the specified minimum and maximum speeds, respectively.

Given an overall solution {$\langle c,p \rangle$},
the combined profit is {\small $P(p) = \sum_{i=1}^{m}p_i\pi_i$},
the travelling time to city {$c_k$} is {\small $T_{c,p}(k) =  \sum_{k'=1}^{k-1}d(c_{k'},c_{k'+1})/v_{c,p}(k')$},
and the combined travelling time is {\small $T(c,p) = T_{c,p}(n+1) = T_{c,p}(n)+d(c_n,c_1)/v_{c,p}(n)$}.
The goal of a TTP solution is to maximise the overall gain by maximising the overall profit
while at the same time minimising the overall renting cost of the knapsack.
More formally, the goal is to maximise the {\it objective function} $G(c,p)$, defined as

\noindent
\begin{equation}
  G(c,p) = P(p) - R \times T(c,p)
  \label{eqn:ttp_objective_function}
\end{equation}

\noindent
for any possible configuration of {$c$} (cyclic tour) and {$p$} (item collection plan).

In a similar manner to the co-operational co-evolution approach~\cite{potter1994cooperative},
where a problem is divided to several sub-problems and each sub-problem is solved by a separate module,
TTP is often decomposed to its KP and TSP components~\cite{bonyadi2014socially},
with each component solved by a dedicated solver.
In solving the TSP component, the collection plan $p$ and hence {$W_p(i)$} for all cities {$1 \leq i \leq n$
are considered fixed;
the aim is to minimise the combined travelling time {$T(c,p)$} over any possible cyclic tour {$c$}. 
In solving the KP component, the cyclic tour {$c$} and hence {$c_k$} for all positions {$1 \leq k \leq n$} are considered fixed;
the aim is to maximise {$G(c,p)$} over any possible collection plan {$p$}.
To generate an initial solution,
either a cyclic tour {$c$} is generated by assuming an empty collection plan {$p$}~\cite{el2018efficiently} (where no item is considered collected),
or a collection plan {$p$} is generated assuming that all distances between cities are zero at the start~\cite{bonyadi2014socially}.

To iteratively solve the TSP component, a segment reversal heuristic known as \mbox{2-OPT}~\cite{croes1958method} is often used.
The underlying \mbox{{2-OPT($c,k',k''$)}} function is defined as follows.
Given a tour {$c$} as well as positions {$k'$} and {$k''$} (under the constraint of {$1 < k' < k'' \leq n$}),
the order of the visited cities between {$k'$} and {$k''$} is reversed to obtain a new tour~{$c'$}.
Here {$c'_{k'+k} = c_{k''-k}$} is obtained where {$0 \leq k \leq k'' - k'$}.

To iteratively solve the KP component, the \mbox{bit-flip} operator is often used~\cite{polyakovskiy2014comprehensive,faulkner2015approximate}.
We define the flipping function as \mbox{Flip($p,j$)},
where given a collection plan~{$p$} and an item~{$j$},
the collection state~{$p_j$} is flipped from~{$0$} to~{$1$} or vice versa to obtain a new collection plan~{$p'$}.

To evaluate the effects of each application of \mbox{{2-OPT($c,k',k''$)}} and \mbox{Flip($p,j$)},
the corresponding objective functions {$G(c',p)$} and {$G(c,p')$} must be recalculated. 
This necessitates recalculating {$W_{c',p}(k)$} and {$T_{c',p}(k+1)$}
for all positions {$k' \leq k \leq k''$} in \mbox{{2-OPT($c,k',k''$)}},
as well as {$W_{c,p'}(k)$} and {$T_{c,p'}(k+1)$} 
for all positions {$l_j \leq k \leq n$} in \mbox{Flip($p,j$)}.
This results in an overall computation cost of {$\mathcal{O}(n)$} for {$G(c',p)$} and {$G(c,p')$}.

In the following sections,
we use two functions on sequences of numbers called {\it prefix-minimum} and {\it postfix-maximum}.
For any position $k$ of a sequence of $n$ numbers $S=\langle S(1), S(2), ..., S(n)\rangle$,
the {\it prefix-minimum} function is defined as:

\noindent
\begin{equation}
  \Pi(S,k) = \min(\Pi(S,k-1),S(k))
  \label{eqn:prefix_minimum}
\end{equation}

\noindent
where $\Pi(S,1) = S(1)$.
The {\it postfix-maximum} function is defined as:

\noindent
\begin{equation}
  \Omega(S,k) = \max(S(k),\Omega(S,k+1))
  \label{eqn:prefix_maximum}
\end{equation}

\noindent
where $\Omega(S,n) = S(n)$.
In other words, the prefix-minimum function returns the smallest number among the first $k$ numbers,
while the postfix-maximum function returns the largest number among the last $n-k+1$ numbers
for each position $k$ in the sequence of numbers $S$. 
For example, consider the sequence $S=\langle 9, 6, 8, 4, 5, 7\rangle$.
The corresponding sequences generated via the prefix-minimum and postfix-maximum functions are
$\Pi(S)=\langle 9, 6, 6, 4, 4, 4\rangle$
and
$\Omega(S)=\langle 9, 8, 8, 7, 7, 7\rangle$,
respectively.

\section{Profit Guided Coordination Heuristic}
\label{sec:pgch}

In the definition of TTP given in Section~\ref{sec:background}, a plan for collecting the items is required.
The items are scattered across the cities,
with the collection order necessarily limited by the order of the cities in the cyclic tour.
This suggests that in a TTP solution, monotonous ordering of item collection should not be expected.
Constructive methods such as PackIterative~\cite{faulkner2015approximate} and Insertion~\cite{mei2014improving}
use profitability ratios of the items in conjunction with the distances of the respective cities from the end of the tour.
However, in iterative methods which change the order of the 
cities in solving the TSP component, corresponding changes are required to the 
collection plan. 

As mentioned in Section~\ref{sec:background}, the 2-OPT segment reversal heuristic is often employed for solving the TSP component.
As the length of the segment to be reversed increases,
the amount of corresponding changes required for the collection plan is likely to increase.
Within the context of a meta-optimiser that interleaves solving the KP and TSP components,
the changes to the collection plan are postponed until the dedicated KP solver is executed. 
However, if a reversed segment is not accepted while solving the TSP component,
there is no opportunity to evaluate corresponding changes to the collection plan for the KP component.
This unnecessarily restricts the search space,
as potentially beneficial combinations of segment reversal with corresponding changes to the collection plan are not even attempted.

As an example, consider the simple TTP instance shown in \figurename~\ref{SampleTTP},
which has {$n=5$} cities and {$m=4$} items.
Suppose that the capacity of knapsack {$W=6$},
maximum speed {$v_{max}=1$}, minimum speed {$v_{min}=0.1$}
and the renting rate of the knapsack {$R=1$}.
Furthermore, suppose that an interim solution
has the cyclic tour \mbox{$c=[1,2,3,4,5]$}
and the collection plan \mbox{$p=[0,0,1,1]$} (ie., items {$3$} and {$4$} are collected).
The resultant objective value (see Eqn.~(\ref{eqn:ttp_objective_function})) for this solution is \mbox{$G(c,p) = 4$}.
Using the 2-OPT heuristic for reversing the segment {$[2,3,4]$} in the cyclic tour {$c$},
we obtain the candidate cyclic tour \mbox{$c'=[1,4,3,2,5]$}.
Without changing the collection plan {$p$},
the corresponding objective value is \mbox{$G(c',p) = -1.5$},
which can result in the rejection of the tour modification.
However, if the collection plan is fortuitously changed to {$p'=[1,0,0,1]$},
where item~{$1$} is collected and item~{$3$} is uncollected,
the resultant objective value (gain) is \mbox{$G(c',p')=6$}.
Hence changing the collection plan in coordination with reversing the segment can result in a higher overall gain.

\begin{figure}[!b]
  \centering
  \includegraphics[width=0.55\textwidth]{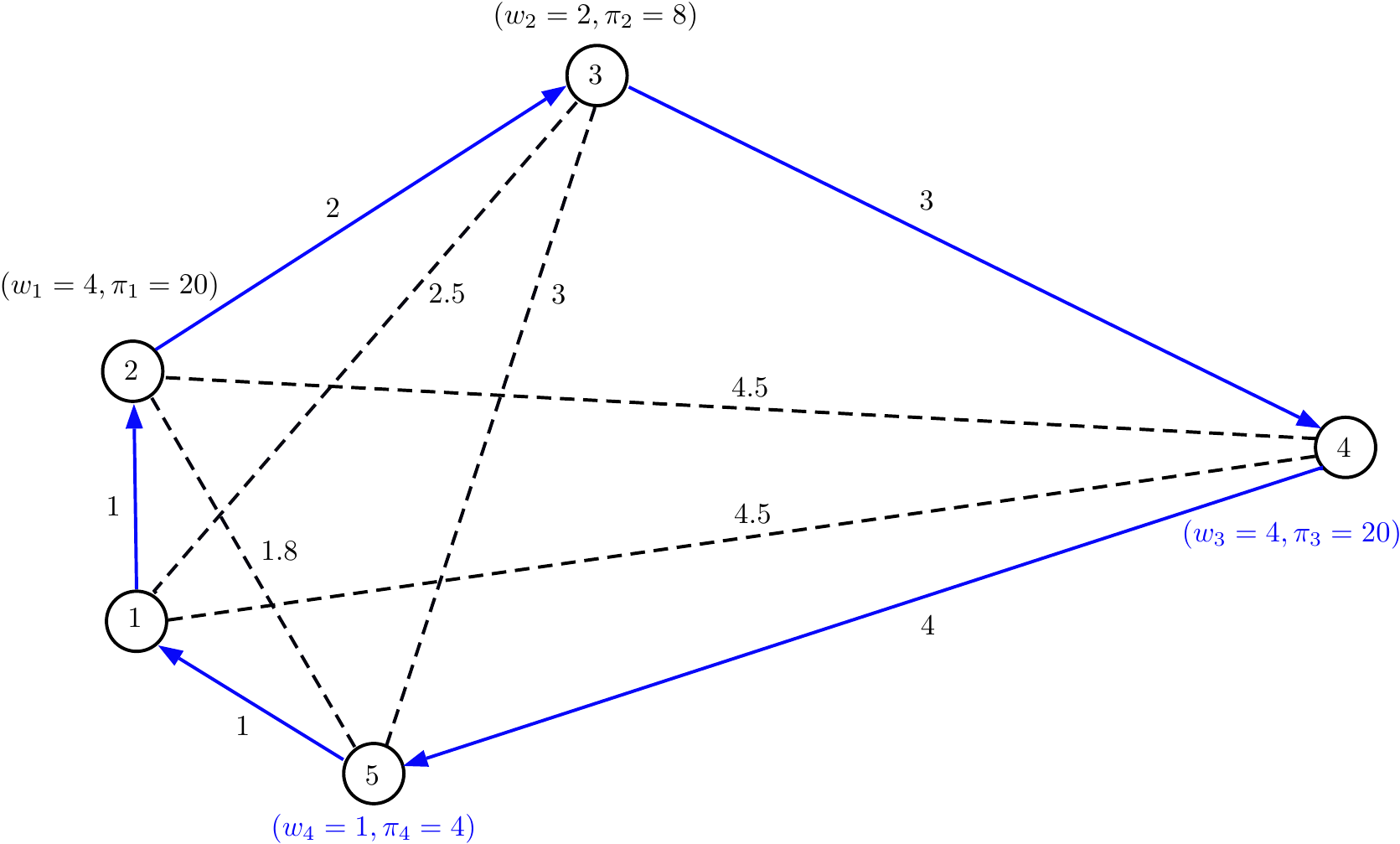}
  \caption{An example TTP instance with 5 cities and 4 items.}
  \label{SampleTTP}
\end{figure}

To see how to change the collection plan in coordination with segment reversing,
let us first examine the solutions found by the PackIterative method 
(the main building block of the state-of-the-art S5 solver~\cite{faulkner2015approximate}).
Using a solution {$\langle c,p \rangle$},
we consider the lowest profitable item {$p(k)$} collected at each city {$c_k$},
as well as the highest profitable item {$q(k)$} not collected at city {$c_k$}.
We then plot the corresponding sequences of profitability ratios {$P_{c,p}(k) = r_{p(k)}$} and {$Q_{c,p}(k) = r_{q(k)}$}, respectively.
If no items are collected at city {$c_k$},
we use the default maximum value of {$P_{c,p}(k) = 1+\max_i r_i$},
where {\it $\max_i r_i$} is the maximum $r$ among all the items.
Furthermore, if there are no uncollected items at city {$c_k$},
we use default minimum value of {$Q_{c,p}(k) = 0$}.

\figurename~\ref{Eil76_1}(a) presents the lowest collected and the highest uncollected profitability ratios
obtained for the {\it eil76\_n750\_uncorr\_10.ttp} benchmark instance
(the benchmark instances are described in Section \ref{sec:experiments}).
Looking {\it forwards} (from the start to the end of the cyclic tour),
the discernible trend is a decrease in the lowest collected profitability ratios in the $P_{c,p}$ sequence.
Furthermore, looking {\it backwards} (from the end to the start of the cyclic tour),
the discernible trend is an increase in the highest uncollected profitability ratios in the $Q_{c,p}$ sequence.

\begin{figure*}[t]
  \centering
  \begin{minipage}{0.49\textwidth}
    \centering
    \includegraphics[width=\textwidth]{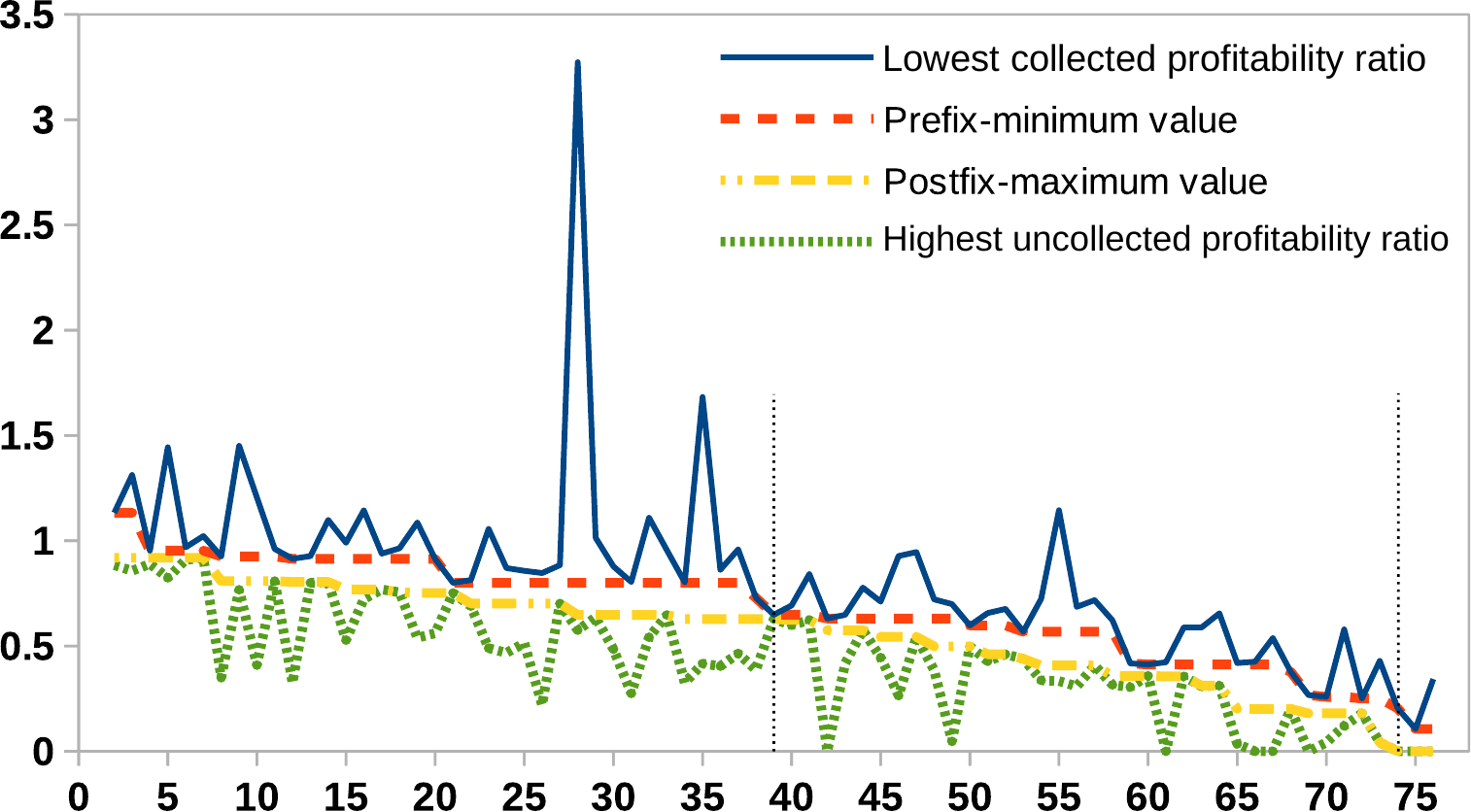}\\
    {\small\bf (a)}
  \end{minipage}
  \hfill
  \begin{minipage}{0.49\textwidth}
    \centering
    \includegraphics[width=\textwidth]{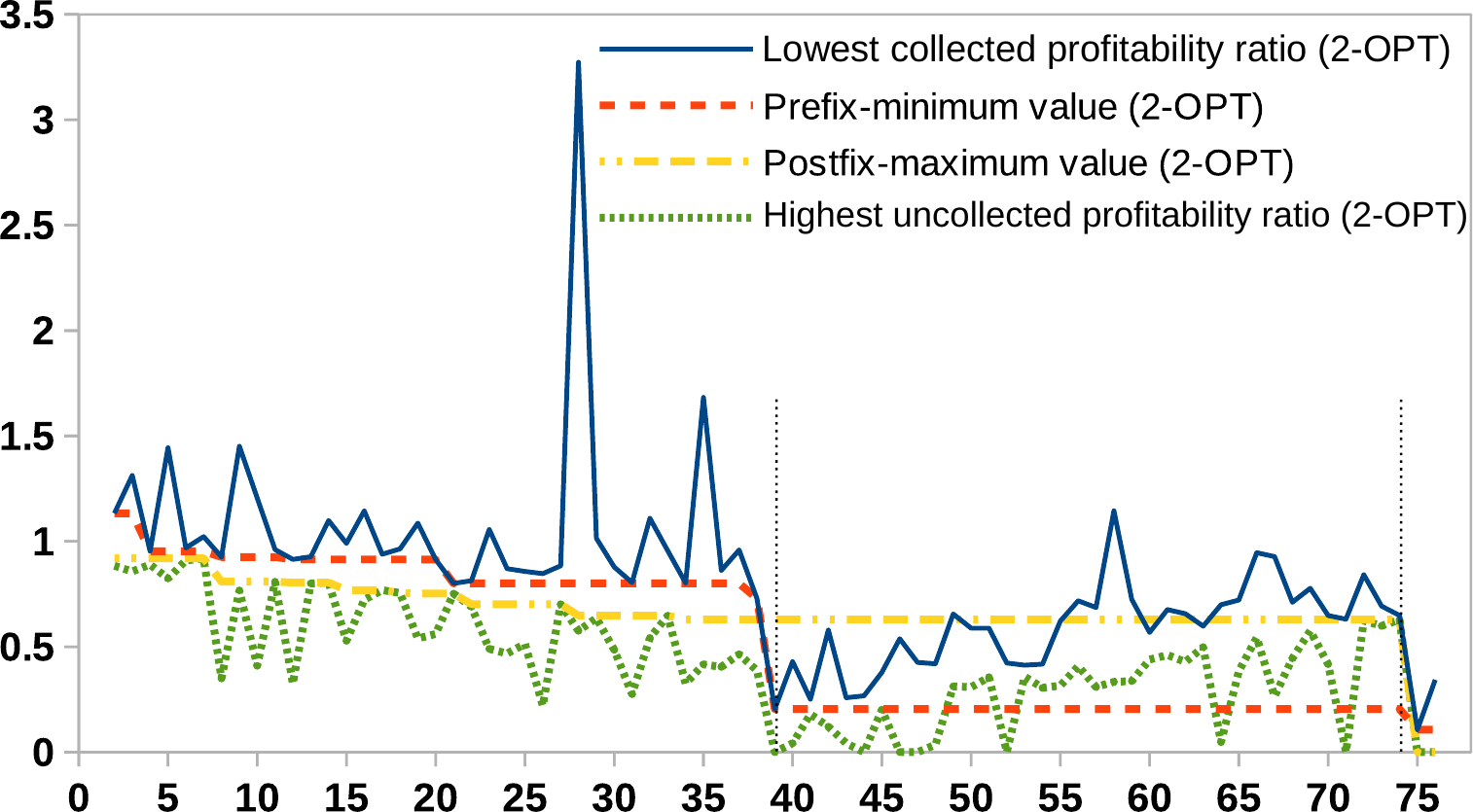}\\
    {\small\bf (b)}
  \end{minipage}
  \caption
    {
    \small
    {\it x}-axis: position in a tour;
    {\it y}-axis: profitability ratio.
    {\bf (a)}: Lowest collected and highest uncollected profitability ratios with the correspondent prefix-minimum and postfix-maximum values
    in a TTP solution obtained by the PackIterative method for the {\it eil76\_n750\_uncorr\_10.ttp} instance, where the overall gain is {$77544.88$}.
    {\bf (b)}: Consequence of employing the 2-OPT heuristic, where the segment between positions 39 and 74 is reversed and the collection plan is unchanged,
    resulting in a lower overall gain of {$72151.46$}.
    }
  \label{Eil76_1}
\end{figure*}

To describe the declining trend of the lowest collected profitability ratios in the $P_{c,p}$ sequence looking forwards,
we use the prefix-minimum function $\Pi(P_{c,p},k)$ as defined in Eqn.~(\ref{eqn:prefix_minimum}),
which provides the profitability ratio of the lowest profitable item collected among the first {$k$} cities.
Furthermore, to describe the rising trend of the highest uncollected profitability ratios in the $Q_{c,p}$ sequence looking backwards,
we use the postfix-maximum function $\Omega(Q_{c,p},k)$ as defined in Eqn.~(\ref{eqn:prefix_maximum}),
which provides the profitability ratio of the highest profitable item not collected among the last {$n-k+1$} cities.
\figurename~\ref{Eil76_1}(a) presents the prefix-minimum and postfix-maximum values
corresponding to the lowest collected and the highest uncollected profitability ratios, respectively. 

Let us employ the \mbox{2-OPT} heuristic
on a segment between positions 39 and 74 of the cyclic tour given in \figurename~\ref{Eil76_1}(a).
The cities between the two positions are reversed,
resulting in the tour presented in \figurename~\ref{Eil76_1}(b).
The reversal also results in lower prefix-minimum values and higher postfix-maximum values 
than the corresponding original prefix-minimum and postfix-maximum values at most positions in the segment.
In comparison to the original segment (as well as the entire cyclic tour),
in the reversed segment the trend of the lowest collected profitability ratios is rising (looking forwards),
and the trend of the highest uncollected profitability ratios is declining (looking backwards).
This reversal in trends makes such a 2-OPT move counterproductive and results in a lower overall gain.
This may then lead to the move being rejected by a solver due to the lower overall gain.
Since the collection plan is not modified, the trade-off between the renting cost and profit is meager
for the low profitable items collected from the start of the reversed segment.

We propose to minimise the renting cost of the knapsack and maximise the profit of the collected items
by adjusting the collection plan in coordination with reversal of the segment.
To achieve this, we propose an extended form of the \mbox{2-OPT} heuristic,
termed as Profit Guided Coordination Heuristic (PGCH).
Along with reversing the tour segment,
items regarded as less profitable and collected in cities located earlier in the reversed segment are first uncollected.
The original prefix-minimum values at the specified tour positions are used 
as a reference to determine which items must be uncollected.
Then, items that tend to be equally or more profitable and not collected in cities located later in the reversed segment are collected.
The original postfix-maximum values at the specified tour positions
are used as an anchor to determine which items can be collected.

\begin{figure*}[!tb]
  \centering
  \begin{subfigure}{0.49\textwidth}
    \centering
    \includegraphics[width=\textwidth]{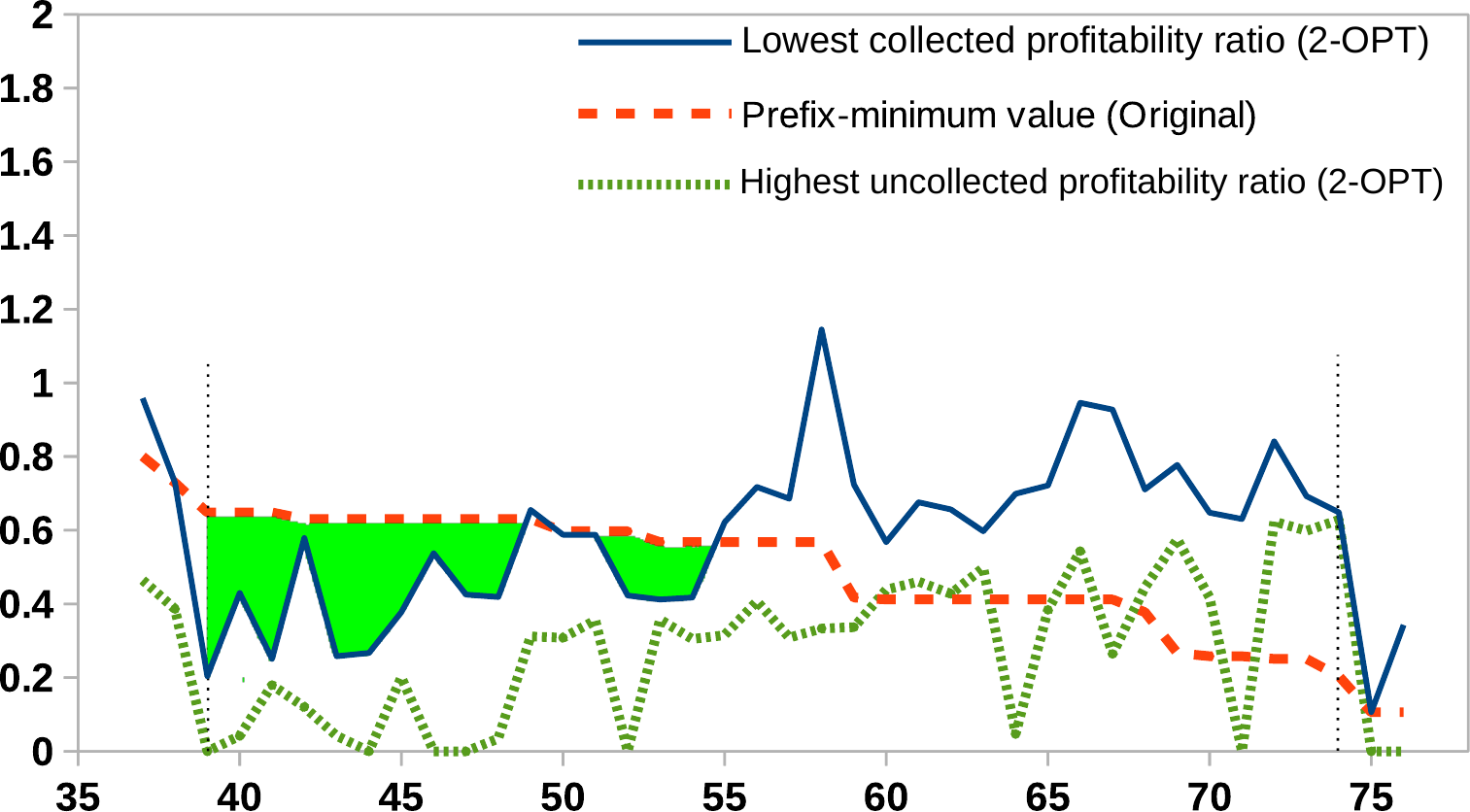} \\
    {\small\bf (a)}
  \end{subfigure}
  \hfill
  \begin{subfigure}{0.49\textwidth}
    \centering
    \includegraphics[width=\textwidth]{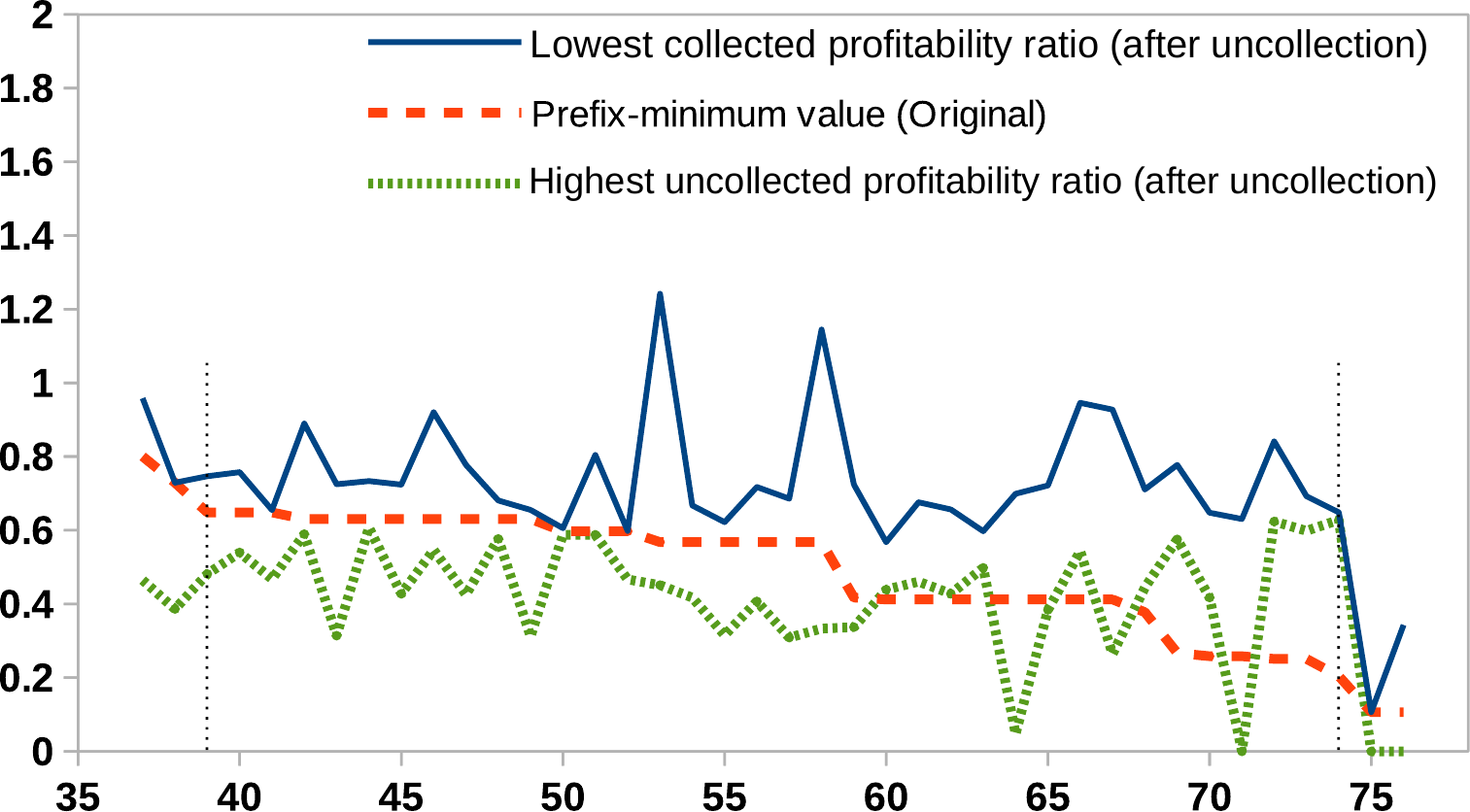}\\
    {\small\bf (b)}
  \end{subfigure}
  \caption
    {
    \small
    {\it x}-axis: position in a tour;
    {\it y}-axis: profitability ratio.
    {\bf (a)}: Lowest collected and highest uncollected profitability ratios of the reversed segment from \figurename~\ref{Eil76_1}(b),
    with an overlay of original prefix-minimum values (dashed red line) from \figurename~\ref{Eil76_1}(a).
    Highlighted green regions under the dashed red line and above the solid blue line indicate the items that must be uncollected.
    {\bf (b)}: Lowest collected and highest uncollected profitability ratios after uncollecting the items as indicated in (a).
    }
  \label{Eil76_2}

\end{figure*}

\begin{figure*}[!tb]
  \centering
  \begin{subfigure}{0.49\textwidth}
    \centering
    \includegraphics[width=\textwidth]{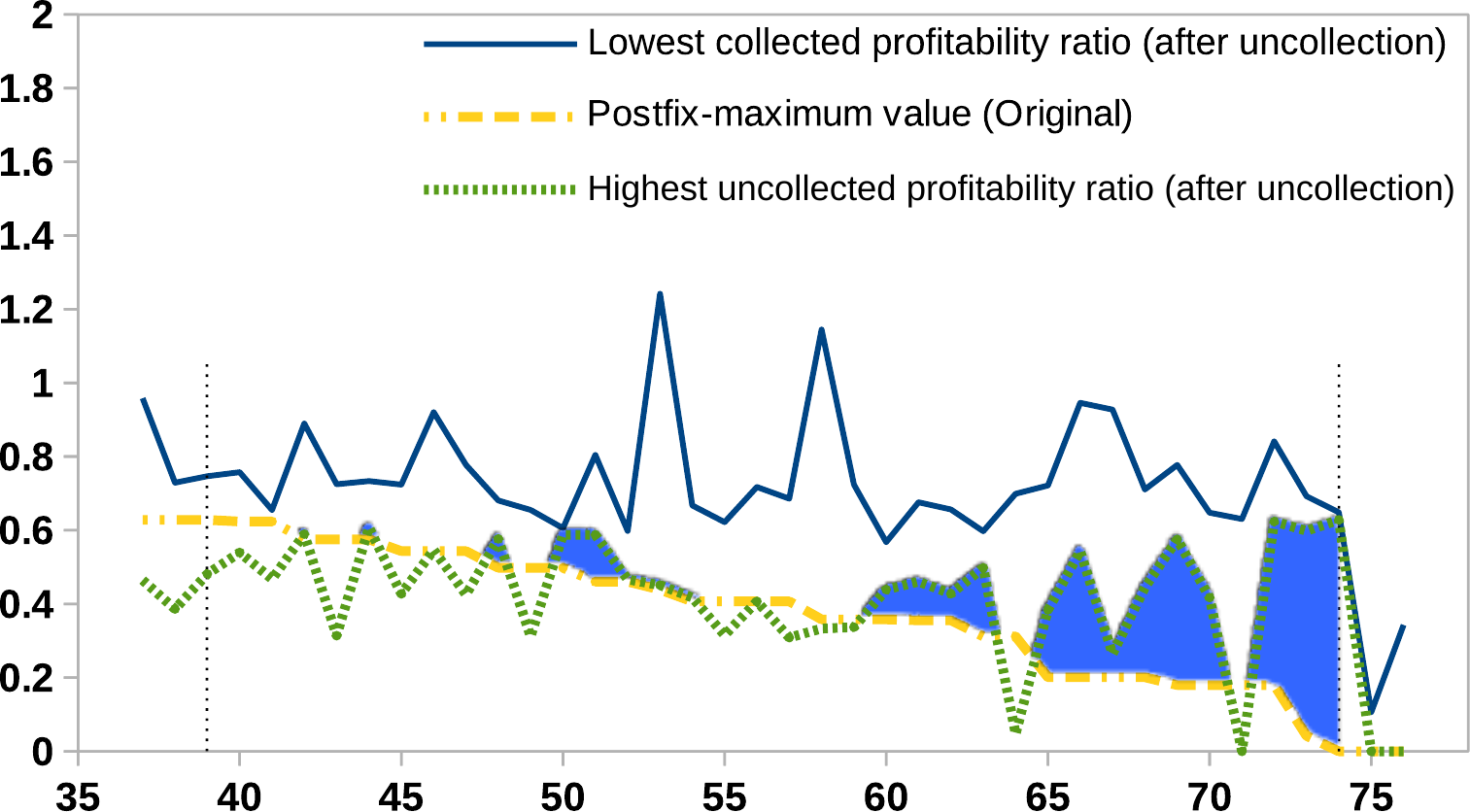} \\
    {\small\bf (a)}
  \end{subfigure}
  \hfill
  \begin{subfigure}{0.49\textwidth}
    \centering
    \includegraphics[width=\textwidth]{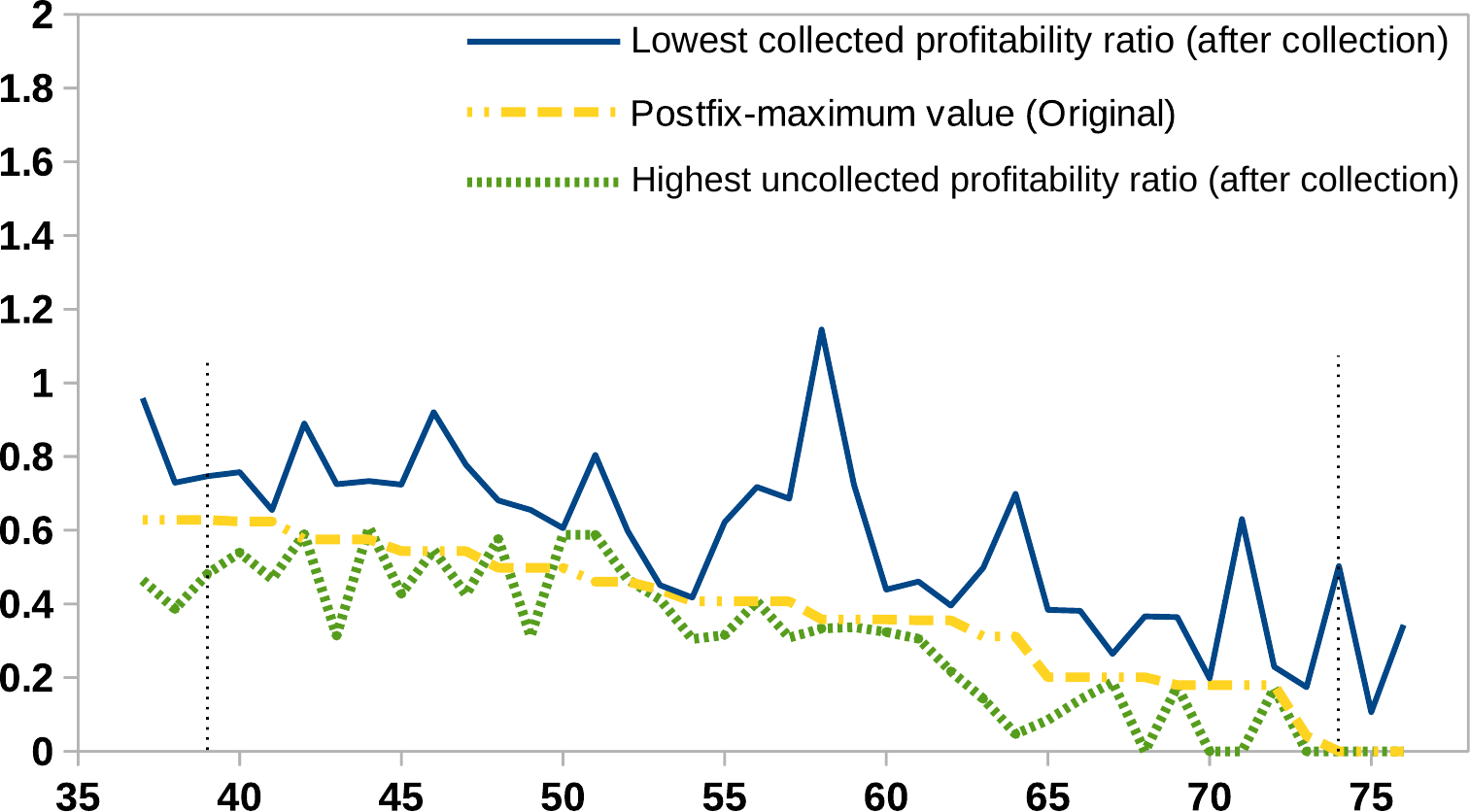}\\
    {\small\bf (b)}
  \end{subfigure}
  \caption
    {
    \small
    {\it x}-axis: position in a tour;
    {\it y}-axis: profitability ratio.
    {\bf (a)}: Lowest collected and highest uncollected profitability ratios from \figurename~\ref{Eil76_2}(b),
    with an overlay of original postfix-maximum values (dashed yellow line) from \figurename~\ref{Eil76_1}(a).
    Highlighted blue regions under the dotted green line and above the dotted-dashed yellow line indicate the items which can be collected.
    {\bf (b)}: Lowest collected and highest uncollected profitability ratios after collecting most of the items as indicated in (a).
    }
  \label{Eil76_3}
\end{figure*}

\begin{figure*}[!tb]
  \centering
  \begin{minipage}{0.49\textwidth}
    \centering
    \includegraphics[width=\textwidth]{eil76DOPT}\\
    {\small\bf (a)}
  \end{minipage}
  \hfill
  \begin{minipage}{0.49\textwidth}
    \centering
    \includegraphics[width=\textwidth]{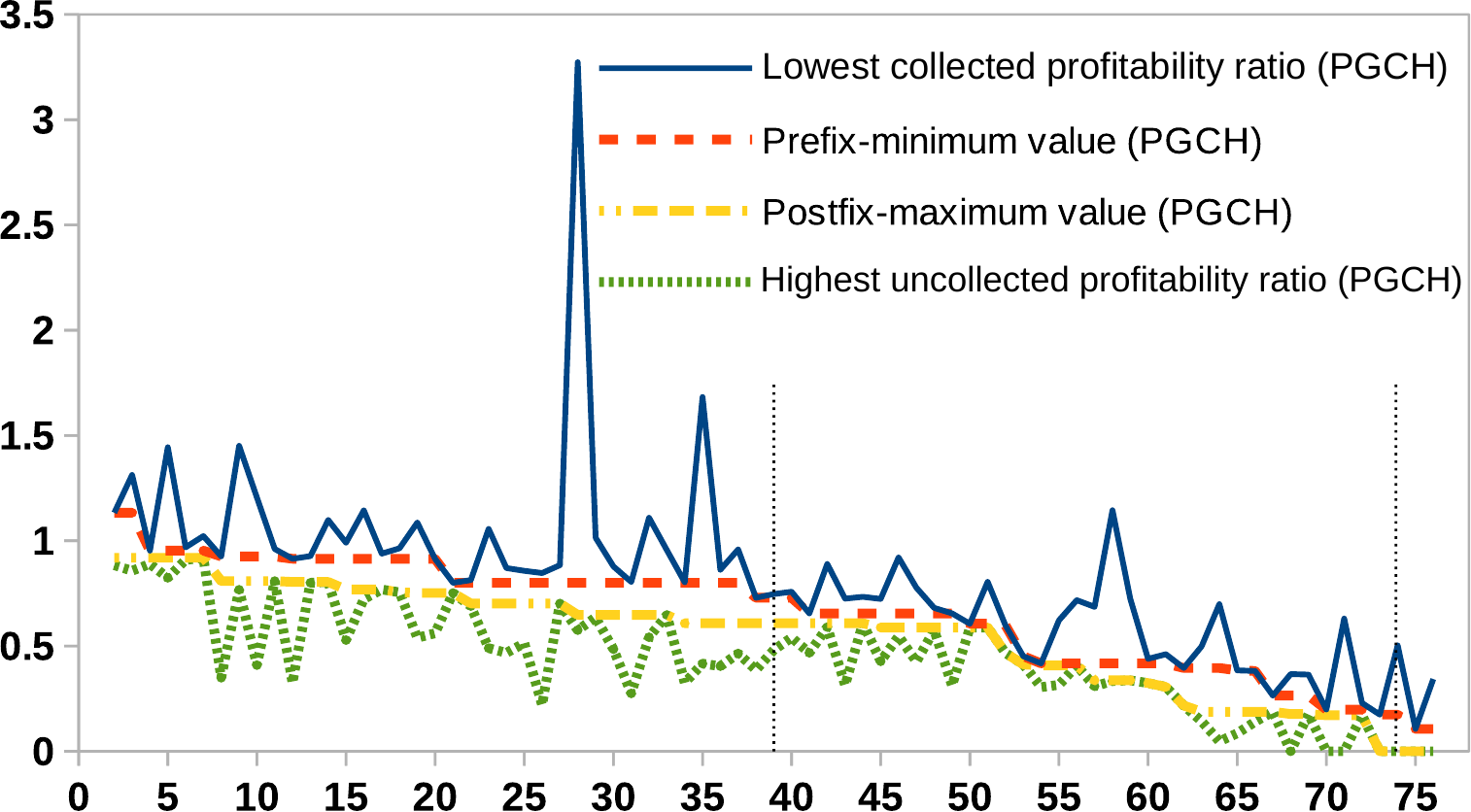}\\
    {\small\bf (b)}
  \end{minipage}
  \caption
    {
    \small
    {\it x}-axis: position in a tour;
    {\it y}-axis: profitability ratio.
    {\bf (a)}:~Copy of \figurename~\ref{Eil76_1}(b), where the 2-OPT heuristic is employed on the segment between positions 39 and 74 of \figurename~\ref{Eil76_1}(a),
    resulting in a overall gain of {$72151.46$}.
    {\bf (b)}:~Consequence of employing PGCH instead of 2-OPT, resulting in an improved overall gain of {$78525.18$}.
    }
  \label{Eil76_4}
\end{figure*}

\clearpage

For example, \figurename~\ref{Eil76_2}(a) shows the lowest collected and the highest uncollected profitability ratios after reversing the segment from \figurename~\ref{Eil76_1}(b),
with an overlay of original prefix-minimum values from \figurename~\ref{Eil76_1}(a).
The highlighted green regions show items which must be uncollected from early positions of the reversed segment.
The result of uncollecting the required items is shown in \figurename~\ref{Eil76_2}(b).
\figurename~\ref{Eil76_3}(a) shows the lowest collected and highest uncollected profitability ratios from \figurename~\ref{Eil76_2}(b),
with an overlay of original postfix-maximum values from \figurename~\ref{Eil76_1}(a).
Highlighted blue regions
show items which can be collected from later positions of the reversed segment. 
The result of collecting most of the items is shown in \figurename~\ref{Eil76_3}(b).
\figurename~\ref{Eil76_4} contrasts the effects of the 2-OPT and PGCH moves
applied to the segment between positions 39 and 74 in \figurename~\ref{Eil76_1}(a).
The tour modified by PGCH obtains a larger overall gain than both the original tour
and the tour modified by 2-OPT.

The formal definition of \textbf{PGCH($c,p,k',k''$)} used in the preceding example is as follows.
Given the positions {$k'$} and {$k''$} (where {$1 < k' < k'' \leq n$}) as well as the solution {$\langle c,p \rangle$},
a candidate solution {$\langle c',p'\rangle$} is generated.
At the start, $p'$ is set to $p$ and the tour $c'$ is obtained
such that {$c'_{k'+k} = c_{k''-k}$} for {$0 \leq k \leq k'' - k'$}.
Then, for each {$k'\leq k \leq k''$},
each item {$j: p'_j = 1$} from city {$c'_k = l_j$} is uncollected (ie.,~{$p'_j$} is set to {$0$}) if {$r_j < \Pi(P_{c,p},k)$.
Moreover, for each {$k''\geq k\geq k'$},
each item {$j : p'_j = 0$} from city {$c'_k = l_j$} is collected (ie.,~{$p'_j$} is set to {$1$}) if {$r_j > \Omega(Q_{c,p},k)$},
as long as the combined weight of the newly collected items
is not larger than the combined weight of the newly uncollected items in the reversed segment.
If no items are uncollected and substituted by other items, PGCH acts like a typical \mbox{2-OPT} heuristic.

To evaluate the effects of each application of PGCH,
data required by the objective function needs to be updated.
Updating {$W_{p'}(c'_k)$} for all positions {$k' \leq k \leq k''$}
as well as
{$W_{c',p'}(k)$} and {$T_{c',p'}(k+1)$} for all positions {$k' \leq k \leq n$}
results in the cost of {$\mathcal{O}(\max(n,m))$} to compute the overall gain.

\section{Boundary Bit-Flip Search}
\label{sec:boundary}

A hill-climber known as bit-flip search has been previously used
for solving the KP component within the TTP setting~\cite{polyakovskiy2014comprehensive,faulkner2015approximate}.
The collection status is flipped (from collected to uncollected and vice-versa) of one item at a time.
A~change to the collection plan is kept if it improves the objective value.
While this is straightforward, the downside is that each change to the collection plan is small and untargeted,
resulting in a slow and meandering exploration of the solution space.
When a time limit is placed for finding a solution,
the solution space may not be explored adequately,
which can contribute to low quality TTP solutions.

We propose a more targeted form of bit-flip search,
where a restriction is placed to consider changes only to the collection state of {\it boundary items},
which are defined as items whose profitability ratios have notable effect on prefix-minimum and postfix-maximum values.
Our motivation for this restriction is twofold:
(i)~changing the collection status of an item requires re-computation of the time to travel from the location of the item to the end of the cyclic tour,
which can be a time consuming process,
and
(ii)~given a fixed cyclic tour, high quality solutions usually follow the same pattern,
where the lowest collected profitability ratios tend to decline when looking forwards,
while the highest uncollected profitability ratios tend to rise up when looking backwards,
as described in Section~\ref{sec:pgch}.

We formally define the boundary items as follows.
Given a solution {$\langle c,p\rangle$},
the lowest profitable item {$p(k)$} collected in city {$c_k$} is considered as a {\it boundary item}
if it is the lowest profitable item among all the items collected at the first {$k$} cities,
ie., {$\Pi(P_{c,p},k) = P_{c,p}(k) = r_{p(k)}$}.
Furthermore, the highest profitable item {$q(k)$} uncollected in city {$c_k$} is considered as a {\it boundary item}
if it is the highest profitable item among all the items uncollected at the last {$n-k+1$} cities,
ie., {$\Omega(Q_{c,p},k) = Q_{c,p}(k) = r_{q(k)}$}.
Considering any position {$k$}, both the lowest profitable item among all the items collected at the first {$k$} cities
and the highest profitable item among all the items uncollected at the last {$n-k+1$} cities 
are assumed as good candidates to be uncollected and collected, respectively.
When the collection status of a boundary item located in position~{$k$} is flipped, 
the values of {$\Pi(P_{c,p'},k')$} with {$k\leq k'\leq n$}
and the values of {$\Omega(Q_{c,p'},k'')$} with {$1< k''\leq k$}
must be updated.
In both cases, the corresponding set of boundary items needs to be updated as well.
Keeping up to date the {$p(k)$} and {$q(k)$} items in each city {$c_k$},
the cost of updating {$\Pi(P_{c,p'},k')$}, {$\Omega(Q_{c,p'},k'')$},
and the corresponding set of boundary items is {$\mathcal{O}(n)$},
ie., the same as the bit-flip operator.

\clearpage

\begin{algorithm*}[!b]
\footnotesize
\caption
  {
  Cooperative Coordination (CoCo) solver
  employing the proposed Profit Guided Coordination Heuristic (PGCH) and {Boundary Bit-Flip} approaches.
  In TSPSolver(), the number of cities in the {cyclic} tour is specified by $n$,
  while the starting and ending point of the segment to be reversed is specified by $k'$ and $k''$, respectively.
  Each segment has at least two cities.
  The position of city~1 is fixed as the initial position in the cyclic tour.
  }
\label{alg:CoCoSolver}
\adjustbox{valign=t}%
{%
\begin{minipage}{0.26\textwidth}
\begin{algorithmic}
\STATE \textbf{proc} CoCoSolver()
\STATE $\langle c_\ast,p_*\rangle \gets \emptyset$ ~ \COMMENT{best solution}
\STATE \textbf{while not global-timeout do}
\STATE \quad$c\gets\mbox{ChainedLKTour}()$
\STATE \quad$p\gets \mbox{InitCollectionPlan}(c)$
\STATE \quad\textbf{while not global-timeout do} 
\STATE \quad\quad$\langle c,p\rangle \gets \mbox{TSPSolver}(c,p)$
\STATE \quad\quad $G_\text{TSP}\gets G(c,p)$
\STATE \quad\quad$p \gets \mbox{KPSolver}(c,p)$
\STATE \quad\quad\textbf{if} $G(c,p)=G_\text{TSP}$
\STATE \quad\quad\quad\textbf{break}  ~ \COMMENT{escape inner loop}
\STATE \quad\quad\textbf{end if}
\STATE \quad\textbf{end while}
\STATE \quad\textbf{if} $G(c,p) > G(c_*,p_*)$
\STATE \quad\quad$\langle c_*,p_*\rangle \gets \langle c,p\rangle$
\STATE \quad\textbf{end if}
\STATE \textbf{end while}
\STATE \textbf{return} {$\langle c_*, p_*\rangle$}
\end{algorithmic}%
\end{minipage}%
}
\adjustbox{valign=t}%
{%
\begin{minipage}[!t]{0.46\textwidth}
\begin{algorithmic}
\STATE \textbf{proc} TSPSolver($c,p$)
\STATE $\langle c_\ddagger,p_\ddagger\rangle \gets \langle c,p\rangle$ ~ {\COMMENT{best candidate solution}}
\STATE \textbf{repeat}
\STATE \quad{$G_\textrm{prev}\leftarrow G(c,p)$}
\STATE \quad\textbf{for} $k' \leftarrow 2$ \textbf{to} $n-1$ \textbf{do} 
\STATE \quad\quad\textbf{foreach} $c_{k''} \in \mbox{DelaTriNeighb}[c_{k'}]$ \textbf{with} $k' < k'' \leq n$
\STATE \quad\quad\quad$\langle c',p'\rangle \gets \mbox{PGCH}(c,p,k',k'')$
\STATE \quad\quad\quad\textbf{if} $G(c',p')>G(c_\ddagger,p_\ddagger)$~\textbf{then}
\STATE \quad\quad\quad\quad$\langle c_\ddagger,p_\ddagger\rangle \gets \langle c',p'\rangle$ \COMMENT{new best candidate solution}
\STATE \quad\quad\quad\textbf{end if}
\STATE \quad\quad\textbf{end foreach}
\STATE \quad\textbf{end for}
\STATE \quad $\langle c,p\rangle \gets \langle c_\ddagger,p_\ddagger\rangle$ \COMMENT{only the best PGCH move takes effect}
\STATE {\textbf{while} $G(c,p)-G_\textrm{prev}\geq \alpha \cdot G_\textrm{prev}$ ~~~~\COMMENT {$\alpha = 0.1 \times 10^{-3}$}}
\STATE \textbf{return} $\langle c, p\rangle$
\end{algorithmic}%
\end{minipage}%
}
\adjustbox{valign=t}%
{%
\begin{minipage}{0.25\textwidth}
\begin{algorithmic}
\STATE \textbf{proc} KPSolver($c,p$)
\STATE $b \gets \mbox{BoundaryItems}(c,p)$
\STATE $ \mbox{MarkUnCheckedAll(b)}$
\STATE \textbf{while not} $\mbox{AllChecked}(b)$ \textbf{do}
\STATE \quad $j \gets \mbox{RandUnCheckedItem}(b)$
\STATE \quad $ \mbox {MarkChecked}(j)$
\STATE \quad $p' \gets \mbox {{Flip}(p,j)}$
\STATE \quad \textbf{if} $G(c,p') > G(c,p)$~\textbf{then}
\STATE \quad\quad$ p \gets p'$
\STATE \quad\quad$ \mbox{Update}(b)$
\STATE \quad\quad$ \mbox{MarkUnCheckedAll}(b)$
\STATE \quad\textbf{end if}
\STATE \textbf{end while}
\STATE \textbf{return} $p$
\end{algorithmic}%
\end{minipage}%
}
\end{algorithm*}

\section{Cooperative Coordination Solver}
\label{sec:solver}

The proposed \mbox{PGCH} and Boundary Bit-flip approaches are employed in the Cooperative Coordination (CoCo) solver presented in Algorithm \ref{alg:CoCoSolver}.
Within the \mbox{CoCoSolver()} function,
an initial {cyclic} tour is generated using the oft-used Chained Lin-Kernighan heuristic~\cite{applegate2003chained}
via the \mbox{ChainedLKTour()} function.
The \mbox{InitCollectionPlan()} function provides an initial collection plan,
which is the plan with the largest TTP overall gain from the plans provided 
by the PackIterative~\cite{faulkner2015approximate} and Insertion~\cite{mei2014improving} methods.
The initial overall TTP solution is then iteratively refined
through the \mbox{TSPSolver()} and \mbox{KPSolver()} functions in an interleaved form.
If in any iteration the solution provided by \mbox{TSPSolver()}
is not improved by \mbox{KPSolver()} in terms of the overall gain value,
the refining process of the current solution is aborted.
The entire process is iteratively restarted while staying within a specified time budget.

The \mbox{TSPSolver()} function is a steepest ascent hill-climbing method which acts as follows.
Given a solution {$\langle c, p \rangle$},
the best candidate solution {$\langle c_\ddagger, p_\ddagger \rangle$} is first considered to be the same as {$\langle c, p \rangle$}.
Then, for each position {$k'$} with {$1<k'<n$},
the pre-computed Delaunay triangulation~\cite{delaunay1934sphare} neighbourhood for city~{$c_{k'}$} is taken into account,
as done by other TSP solvers~\cite{el2018efficiently}.
For each city \mbox{$c_{k''} : k'$ $<$ $k''$ $\leq$ $n$} within the neighbourhood indicated by $\mbox{DelaTriNeighb}[c_{k'}]$,
the proposed \mbox{PGCH} approach is employed to obtain a new candidate solution {$\langle c', p'\rangle$}.
If the new candidate solution has a larger overall gain than the best candidate solution found so far,
it is accepted as the new best candidate solution.
After examining all positions in the tour,
the current solution {$\langle c, p \rangle$} is substituted by the best candidate solution {$\langle c_\ddagger, p_\ddagger \rangle$}.
Hence only the best \mbox{PGCH} move takes effect, which modifies both $c$ and $p$.
If the current solution has a overall gain that has sufficiently changed
(with an empirically quantified margin of {$\alpha = 0.01\%$} of the previous solution),
the evaluation of all positions is repeated with the new current solution.
Otherwise, the current solution is returned.

The value of the {$\alpha$} parameter has been empirically set
with the aim to avoid spending time on improvements that are likely to be very minor,
especially when solving large instances. 
Within a given time budget,
this approach allows TSPSolver() to evaluate a larger number of tours,
some possibly more promising, 

In the \mbox{KPSolver()} function,
all boundary items are placed in bag {$b$} via the BoundaryItems$(c,p)$ function,
and are then marked as unchecked by the MarkUnCheckedAll($b$) function.
The AllChecked($b$) function is used to determine whether all items in bag $b$ have been checked.
As long as there is at least one unchecked boundary item in the bag,
a randomly unchecked boundary item $j$ is selected via the RandUnCheckedItem($b$) function.
The selected item is marked as checked via the MarkChecked($j$) function,
and its collection status is flipped (from collected to uncollected or vice versa)
to obtain a new candidate collection plan {$p'$}.
If the new overall gain is larger than the current one, the change is accepted.
In such case, the boundary items bag {$b$} is updated via the Update$(b)$ function,
and all the items inside the bag are marked as unchecked via the MarkUnCheckedAll($b$) function.
The initial filling of the items bag $b$  is done as explained in Section~\ref{sec:boundary}.

\newpage
\section{Experiments}
\label{sec:experiments}

The performance of TTP solvers is usually compared on TTP benchmark instances introduced in~\cite{polyakovskiy2014comprehensive}.
Each TTP benchmark instance has been defined based on: 
\begin{itemize}
\itemsep=0ex
\item A symmetric TSP instance among TSPLIB benchmark instances~\cite{reinelt1991tsplib}, ranging from 51 to 85900 cities.
\item A set of 1, 3, 5 or 10 items per each city indicated by the item-factor parameter.
\item A knapsack-type parameter which indicates that the weights and profits of the items have one of the following characteristics:%
  \begin{itemize}
  \itemsep=0ex
  \item weights and profits are bounded and strongly correlated
  \item weights and profits are uncorrelated, and weights of all items are similar
  \item weights and profits are uncorrelated
  \end{itemize}
\item A knapsack with a capacity level ranging from 1 to 10.
\end{itemize}
For example, the {\it eil76\_n75\_uncorr\_10.ttp} instance discussed in Section~\ref{sec:pgch}
has been defined based on the {\it eil76} symmetric TSP instance including 76 cities,
with just one item per each city (excluding the first city)
where the weights and profits of the items are uncorrelated and a knapsack with the highest capacity level indicated by the number 10. 
 
Four sets of experiments were performed on a comprehensive subset of benchmark instances%
\footnote{The TTP benchmark instances were obtained from {\url{https://cs.adelaide.edu.au/~optlog/CEC2014COMP_InstancesNew/}}} 
placed in 3 categories as per~\cite{el2018efficiently,el2016population}.
We label the 3 categories as A, B, C.
There are 20 instances in each category, ranging from 76 to 33810 cities.
In category~A, knapsack capacity is relatively small; there is only one item in each city; the weights and profits of the items are bounded and strongly correlated.
In category~B, knapsack capacity is moderate; there are 5 items in each city; the weights and profits of the items are uncorrelated; the weights of all items are similar.
In category~C, knapsack capacity is high; there are 10 items in each city; the weights and profits of the items are uncorrelated.

Each solver was independently run on each TTP instance 10 times.
The same experiment setup was used in all experiments, except for necessary modifications in the last experiment.
Specifically, each run had a standard timeout of 10 minutes, excluding the last experiment.
For each run in all experiments,
we ensured that whenever required, each solver computed new initial cyclic tours via the Chained Lin-Kernighan heuristic~\cite{applegate2003chained}.
All experiments were run on a machine with a 2~GB memory limit
and an Intel Xeon CPU X5650 running at 2.66~GHz.

To measure differences in performance across solvers, 
we use the relative deviation index~(RDI)~\cite{kim1996simulated}
for each solver on each TTP instance.
RDI is defined as:

\noindent
\begin{equation}
\textrm{RDI} = (G_{\textrm{mean}}-G_\textrm{min})\times100/(G_\textrm{max}-G_\textrm{min})
\label{eqn:RDI}
\end{equation}

\noindent
where {$G_{\textrm{mean}}$} is the mean of the {$G(c,p)$} overall gain values
of the 10 runs of a solver on an instance,
while {$G_\textrm{min}$} and {$G_\textrm{max}$} are the minimum and the maximum {$G(c,p)$} values, respectively,
obtained by any run of any solver on the same instance in each set of experiments.
For each set of experiments, \ref{sec:appendix} contains corresponding tables 
that show the minimum and maximum {$G(c,p)$} values 
as well as the {$G_{\textrm{mean}}$} value for each solver on each instance.

In the first set of experiments we gauge the outcomes of the proposed PGCH approach and Boundary Bit-Flip search,
as described in Sections~\ref{sec:pgch} and~\ref{sec:boundary}, respectively.
We compare four variants of the TTP solver described in Section~\ref{sec:solver},
denoted as Solver~1, Solver~2, Solver~3, and Solver~4.
For solving the TSP component, either the standard 2-OPT move or the proposed PGCH approach is used.
For solving the KP component, either standard bit-flip search or the proposed Boundary Bit-Flip search is used.
The variants are configured as follows:

\begin{itemize}
\itemsep=0ex
\item Solver~1: 2-OPT + standard bit-flip search
\item Solver~2: 2-OPT + Boundary Bit-Flip search
\item Solver~3:  PGCH + standard bit-flip search
\item Solver~4:  PGCH + Boundary Bit-Flip search (equivalent to the CoCo solver described in Section~\ref{sec:solver})
\end{itemize}

\begin{table*}[!tb]
\centering
\caption
  {
  Performance comparison of four variants of the TTP solver presented in Algorithm~\ref{alg:CoCoSolver},
  reported in terms of the relative deviation index (RDI) defined in Eqn.~(\ref{eqn:RDI}).
  Corresponding average overall gain values are shown in Tables~\ref{tab:exp1catA}, \ref{tab:exp1catB} and \ref{tab:exp1catC} in~\ref{sec:appendix}.
  3~categories of TTP instances are used.
  Category~A:
  knapsack capacity is relatively small;
  1~item in each city;
  weights and profits of items are bounded and strongly correlated.
  Category~B:
  knapsack capacity is moderate;
  5~items in each city;
  weights and profits of items are uncorrelated, and the weights of all items are similar.
  Category~C: 
  knapsack capacity is high;
  10~items in each city;
  weights and profits of items are uncorrelated.
  For solving the TSP component, either the standard 2-OPT move or the proposed PGCH approach is used.
  For solving the KP component, either the standard bit-flip search or the proposed Boundary Bit-Flip search is used.
  The solvers are configured as follows:
  Solver~1~=~2-OPT~+~standard~bit-flip;
  Solver~2~=~2-OPT~+~Boundary~Bit-Flip;
  Solver~3~=~PGCH~+~standard~bit-flip;
  Solver~4~=~PGCH~+~Boundary~Bit-Flip.
  }
\label{tab:experiments_1}
\vspace{-1.5ex}
\small
\begin{tabular}{|l|r|r|r|r||r|r|r|r||r|r|r|r|}
\hline
\multirow{2}{*}{\bf Instance} & \multicolumn{4}{c||}{\textbf{Category A}} & \multicolumn{4}{c||}{\textbf{Category B}} & \multicolumn{4}{c|}{\textbf{Category C}} \\  \cline{2-13}
                                                            & \multicolumn{1}{c|}{\rotatebox[origin=c]{0}{\textbf{\scriptsize Solver~1}}} & \multicolumn{1}{c|}{\rotatebox[origin=c]{0}{\textbf{\scriptsize Solver~2}}} & \multicolumn{1}{c|}{\rotatebox[origin=c]{0}{\textbf{\scriptsize Solver~3}}} & \multicolumn{1}{c||}{\rotatebox[origin=c]{0}{\textbf{\scriptsize Solver~4}}} &  \multicolumn{1}{c|}{\rotatebox[origin=c]{0}{\textbf{\scriptsize Solver~1}}} & \multicolumn{1}{c|}{\rotatebox[origin=c]{0}{\textbf{\scriptsize Solver~2}}} & \multicolumn{1}{c|}{\rotatebox[origin=c]{0}{\textbf{\scriptsize Solver~3}}} & \multicolumn{1}{c||}{\rotatebox[origin=c]{0}{\textbf{\scriptsize Solver~4}}} & \multicolumn{1}{c|}{\rotatebox[origin=c]{0}{\textbf{\scriptsize Solver~1}}} & \multicolumn{1}{c|}{\rotatebox[origin=c]{0}{\textbf{\scriptsize Solver~2}}} & \multicolumn{1}{c|}{\rotatebox[origin=c]{0}{\textbf{\scriptsize Solver~3}}} & \multicolumn{1}{c|}{\rotatebox[origin=c]{0}{\textbf{\scriptsize Solver~4}}} \\ \hline

eil76 & \textbf{100} & \textbf{100} & \textbf{100} & \textbf{100} & 54.1 & 33.8 & \textbf{96.6} & 94.6 & \textbf{65.4} & 56.4 & 53.3&52.0 \\ 
kroA100 & 51.8 & 77.3 & \textbf{95.1} & 84.4 & 1.9 & 2.7 & \textbf{93.2} & 85.3 & 12.1 & 1.1 & \textbf{66.6}&65.9 \\ 
ch130 & 53.2 & 38.0 & \textbf{100} & \textbf{100} & 1.1 & 1.0 & 80.7 & \textbf{99.7} & 51.1 & 46.0 & \textbf{75.6}&30.9 \\ 
u159 & 0.4 & 0.0 & \textbf{60.2} & 54.0 & 20.6 & \textbf{86.6} & 31.8 & 60.1 & 14.8 & 11.4 & \textbf{84.4}&78.4 \\ \hline
a280 & 1.2 & 0.0 & \textbf{75.8} & 65.3 & 0.5 & 2.4 & 99.9 & \textbf{100} & 24.2 & 23.8 & 95.5&\textbf{98.9} \\ 
u574 & 48.3 & 28.8 & 70.2 & \textbf{71.7} & 12.6 & 24.1 & \textbf{82.5} & 79.7 & 18.8 & 25.3 & \textbf{83.9}&77.0 \\ 
u724 & 6.3 & 12.1 & \textbf{70.1} & 68.1 & 6.7 & 8.1 & 83.1 & \textbf{87.6} & 24.6 & 21.9 & 87.9&\textbf{92.4} \\ 
dsj1000 & \textbf{100} & \textbf{100} & \textbf{100} & \textbf{100} & 5.7 & 6.2 & 88.5 & \textbf{89.9} & 21.0 & 35.7 & 47.9&\textbf{76.5} \\ \hline
rl1304 & 26.2 & 30.5 & 46.6 & \textbf{56.4} & 0.4 & 38.9 & 94.1 & \textbf{96.6} & 14.7 & 17.5 & 86.4&\textbf{94.0} \\
fl1577 & 64.2 & 58.4 & \textbf{75.7} & 59.5 & 32.7 & 40.2 & 62.8 & \textbf{74.1} & 29.7 & 57.3 & 62.8&\textbf{77.4} \\
d2103 & 35.7 & 29.1 & \textbf{70.5} & 59.8 & 18.8 & 27.6 & 88.9 & \textbf{95.1} & 11.3 & 16.4 & 69.3&\textbf{76.5} \\
pcb3038 & 20.2 & 16.9 & \textbf{84.1} & 66.1 & 20.2 & 25.6 & 72.3 & \textbf{91.4} & 20.6 & 34.3 & 69.8&\textbf{79.1} \\ \hline
fnl4461 & 19.1 & 23.6 & \textbf{52.1} & 52.0 & 20.7 & 37.4 & 70.4 & \textbf{87.5} & 28.8 & 56.8 & 78.8&\textbf{91.9} \\ 
pla7397 & 53.7 & 54.7 & 72.4 & \textbf{74.0} & 49.0 & 72.5 & 80.4 & \textbf{92.5} & 42.8 & 83.7 & 73.1&\textbf{91.9} \\ 
rl11849 & 16.8 & 15.6 & 50.3 & \textbf{64.8} & 16.8 & 36.8 & 68.4 & \textbf{88.4} & 22.8 & 38.7 & 67.7&\textbf{92.8} \\ 
usa13509 & 34.2 & 36.1 & \textbf{75.8} & 74.9 & 63.9 & 88.0 & 85.2 & \textbf{94.2} & 48.4 & 85.4 & 74.3&\textbf{92.9} \\\hline
brd14051 & 18.2 & 30.6 & \textbf{69.4} & 61.5 & 43.3 & 67.3 & 83.4 & \textbf{93.1} & 48.5 & 83.1 & 92.3&\textbf{96.9} \\ 
d15112 & 24.1 & 19.5 & \textbf{86.8} & 84.5 & 35.6 & 61.3 & 67.3 & \textbf{85.4} & 31.0 & 81.1 & 77.6&\textbf{89.5} \\
d18512 & 36.9 & 20.1 & \textbf{76.0} & 62.3 & 46.1 & 78.0 & 89.4 & \textbf{95.5} & 36.6 & 85.3 & 86.8&\textbf{93.4} \\ 
pla33810 & 34.2 & 31.7 & 65.2 & \textbf{72.7} & 11.9 & 33.3 & 71.6 & \textbf{82.9} & 27.8 & 67.1 & 80.1&\textbf{96.8} \\ \hline\hline
Average & 37.2 & 36.1 & \textbf{74.8} & 71.6 & 23.1 & 38.6 & 79.5 & \textbf{88.7} & 29.8 & 46.4 & 75.7&\textbf{82.3} \\ \hline

\end{tabular}
\end{table*}

The results shown in Table~\ref{tab:experiments_1} indicate that on all three categories,
solvers using the proposed PGCH move (Solver~3 and Solver~4)
perform considerably better than corresponding solvers using the 2-OPT move (Solver~1 and Solver~2).
The results also indicate that on most instances in Categories~B and~C,
solvers using Boundary~Bit-Flip search (Solver~2 and Solver~4),
perform notably better than the corresponding solvers using standard bit-flip search (Solver~1 and Solver~3).
This supports our hypothesis in Section~\ref{sec:boundary}:
restricting the application of the bit-flip operator to the boundary items allows for more efficient utilisation of the time budget
(ie.,~less time spent on the KP component allows more tours to be evaluated),
and favours solutions which follow the pattern of known high quality solutions:
the lowest collected profitability ratios tend to decline when looking forwards,
while the highest uncollected profitability ratios tend to rise up when looking backwards.
In Category~A, Boundary Bit-Flip search leads to somewhat degraded performance compared to standard bit-flip search.
As there is a relative scarcity of items in Category~A compared to B and~C,
restricting the application of the bit-flip operator to the boundary items 
reduces the search space excessively, which in turn can lead to degraded performance.

In the second set of experiments,
we analyse the differences in performance between 2-OPT and PGCH in more detail.
We compare Solver~2 and Solver~4 (the two best performing variants using 2-OPT and PGCH),
based on the average relative length and the average number of the accepted segment reversing moves.
The results presented in Table~\ref{tab:experiments_2} indicate that on average
PGCH leads to notably more accepted moves.
Furthermore, accepted PGCH moves are considerably longer (ie., larger segments).
This supports our hypothesis in Section~\ref{sec:pgch}:
it is better to change the collection plan in direct coordination with segment reversal,
instead of postponing the changes and completely relying on the dedicated KP solver (ie., weak coordination).

\begin{table}[!tb]
\centering
\caption
  {
  Contrasting the effects of the 2-OPT move and the proposed PGCH approach in Solver~2 and Solver 4, respectively.
  The first two columns in each category show the average relative lengths of the accepted segment reversing moves (in \%),
  ie., {$|k''$--~$k'$+$1|\times100/n$}.
  The last two columns in each category show the average number of accepted segment reversing moves.
  }
\label{tab:experiments_2}
\vspace{-1.5ex}
\small
\begin{tabular}{|l|r|r||r|r||r|r||r|r||r|r||r|r|}
\hline
\multirow{3}{*}{\bf Instance} & \multicolumn{4}{c||}{\textbf{Category A}} & \multicolumn{4}{c||}{\textbf{Category B}} & \multicolumn{4}{c|}{\textbf{Category C}} \\  \cline{2-13}

  & \multicolumn{2}{c||}{\textbf{Rel. rev. len. \%}} & \multicolumn{2}{c||}{\textbf{Rev. num}} &
      \multicolumn{2}{c||}{\textbf{Rel. rev. len. \%}} & \multicolumn{2}{c||}{\textbf{Rev. num}} &
      \multicolumn{2}{c||}{\textbf{Rel. rev. len. \%}} & \multicolumn{2}{c|}{\textbf{Rev. num}}
       \\ \cline{2-13}
 
& \multicolumn{1}{c|}{\rotatebox[origin=c]{0}{\textbf{\scriptsize ~Solver~2}}}  & 
    \multicolumn{1}{c||}{\rotatebox[origin=c]{0}{\textbf{\scriptsize Solver~4}}}  &
    \multicolumn{1}{c|}{\rotatebox[origin=c]{0}{\textbf{\scriptsize Solver~2}}}  & 
    \multicolumn{1}{c||}{\rotatebox[origin=c]{0}{\textbf{\scriptsize Solver~4}}}  &
    \multicolumn{1}{c|}{\rotatebox[origin=c]{0}{\textbf{\scriptsize Solver~2}}}  & 
    \multicolumn{1}{c||}{\rotatebox[origin=c]{0}{\textbf{\scriptsize Solver~4}}}  &
    \multicolumn{1}{c|}{\rotatebox[origin=c]{0}{\textbf{\scriptsize Solver~2}}}  & 
    \multicolumn{1}{c||}{\rotatebox[origin=c]{0}{\textbf{\scriptsize Solver~4}}}  &
    \multicolumn{1}{c|}{\rotatebox[origin=c]{0}{\textbf{\scriptsize Solver~2}}}  & 
    \multicolumn{1}{c||}{\rotatebox[origin=c]{0}{\textbf{\scriptsize Solver~4}}}  &
    \multicolumn{1}{c|}{\rotatebox[origin=c]{0}{\textbf{\scriptsize Solver~2}}} & 
    \multicolumn{1}{c|}{\rotatebox[origin=c]{0}{\textbf{\scriptsize Solver~4}}} \\ \hline

eil76 & 49.9 & \textbf{52.2} & 1.6 & \textbf{4.4} & 2.7 & \textbf{54.1} & 0.6 & \textbf{1.2} & 5.8 & \textbf{26.5} & 0.9 & \textbf{2.4} \\ 
kroA100 & 47.0 & \textbf{63.0} & \textbf{1.1} & \textbf{1.1} & 3.2 & \textbf{86.0} & 0.0 & \textbf{0.1} & 2.5 & \textbf{18.3} & 0.0 & \textbf{2.6} \\
ch130 & 7.1 & \textbf{30.2} & \textbf{1.5} & 1.1 & 1.7 & \textbf{48.1} & 0.5 & \textbf{2.3} & 1.6 & \textbf{54.0} & 0.4 & \textbf{1.1} \\
u159 & 2.2 & \textbf{21.9} & 0.4 & \textbf{2.6} & 1.3 & \textbf{21.0} & 2.5 & \textbf{4.4} & 1.3 & \textbf{30.7} & 2.4 & \textbf{4.0} \\ \hline
a280 & 46.8 & \textbf{62.6} & \textbf{4.6} & \textbf{4.6} & 0.8 & \textbf{49.7} & 1.2 & \textbf{5.6} & 0.9 & \textbf{33.8} & 1.3 & \textbf{1.9} \\
u574 & 15.2 & \textbf{28.7} & 3.2 & \textbf{10.5} & 2.5 & \textbf{21.3} & 2.3 & \textbf{8.0} & 2.1 & \textbf{25.7} & 1.7 & \textbf{2.9} \\ 
u724 & 16.9 & \textbf{30.5} & 8.6 & \textbf{12.6} & 2.6 & \textbf{26.3} & 3.6 & \textbf{17.6} & 2.0 & \textbf{37.3} & 2.6 & \textbf{10.2} \\ 
dsj1000 & 0.0 & 0.0 & 0.0 & 0.0 & 5.6 & \textbf{28.6} & 1.3 & \textbf{15.6} & 15.3 & \textbf{26.5} & 2.0 & \textbf{17.4} \\ \hline
rl1304 & 4.8 & \textbf{27.4} & 3.7 & \textbf{8.6} & 2.0 & \textbf{14.3} & 2.6 & \textbf{17.6} & 8.3 & \textbf{21.2} & 2.9 & \textbf{6.3} \\ 
fl1577 & 8.8 & \textbf{25.0} & 9.3 & \textbf{11.9} & 1.9 & \textbf{22.0} & 15.0 & \textbf{31.9} & 6.6 & \textbf{23.7} & 14.5 & \textbf{24.7} \\
d2103 & 10.9 & \textbf{30.2} & 6.8 & \textbf{15.9} & 15.3 & \textbf{51.9} & 4.0 & \textbf{12.3} & 23.2 & \textbf{45.5} & 4.8 & \textbf{11.6} \\ 
pcb3038 & 18.8 & \textbf{32.7} & 18.4 & \textbf{25.5} & 1.0 & \textbf{29.7} & 3.0 & \textbf{33.7} & 0.6 & \textbf{36.1} & 2.8 & \textbf{18.8} \\ \hline
fnl4461 & 12.8 & \textbf{33.7} & 12.9 & \textbf{32.6} & 0.5 & \textbf{20.8} & 5.3 & \textbf{51.4} & 0.5 & \textbf{25.4} & 4.8 & \textbf{33.7} \\ 
pla7397 & 13.6 & \textbf{17.9} & \textbf{78.0} & 76.2 & \textbf{17.9} & \textbf{17.9} & \textbf{113} & 104 & \textbf{20.2} & 19.1 & 92.0 & \textbf{92.7} \\ 
rl11849 & 6.5 & \textbf{32.9} & 26.6 & \textbf{58.0} & 2.2 & \textbf{25.8} & 27.5 & \textbf{114} & 3.1 & \textbf{29.4} & 23.8 & \textbf{86.7} \\ 
usa13509 & 16.8 & \textbf{17.1} & 70.8 & \textbf{83.2} & \textbf{13.1} & 12.3 & 214 & \textbf{293} & \textbf{20.4} & 15.1 & 180 & \textbf{220} \\ \hline
brd14051 & 13.8 & \textbf{24.4} & 66.1 & \textbf{98.2} & 7.9 & \textbf{17.4} & 115 & \textbf{259} & 13.0 & \textbf{19.5} & 123 & \textbf{198} \\ 
d15112 & 12.3 & \textbf{28.3} & 80.7 & \textbf{127} & 10.5 & \textbf{20.1} & 174 & \textbf{265} & 16.1 & \textbf{22.6} & 197 & \textbf{230} \\ 
d18512 & 13.0 & \textbf{23.2} & 73.0 & \textbf{120} & 6.6 & \textbf{18.0} & 121 & \textbf{289} & 12.4 & \textbf{18.7} & 150 & \textbf{210} \\
pla33810 & 6.3 & \textbf{25.2} & 124 & \textbf{165} & 6.9 & \textbf{27.8} & 181 & \textbf{229} & 9.1 & \textbf{29.4} & 166 & \textbf{176} \\ \hline\hline
Average & 16.2 & \textbf{30.3} & 29.5 & \textbf{42.9} & 5.3 & \textbf{30.7} & 49.3 & \textbf{87.7} & 8.2 & \textbf{27.9} & 48.6 & \textbf{67.6} \\ \hline

\end{tabular}
\end{table}

In the third set of experiments, we compare the proposed CoCo solver
(Solver 4 in Table~\ref{tab:experiments_1}) against the following solvers: 
MATLS~\cite{mei2014improving},
S5~\cite{faulkner2015approximate}
and 
CS2SA*~\cite{el2018efficiently}.
The CS2SA* solver was selected due to its recency,
while the MATLS and S5 solvers were selected due to their salient performance reported in~\cite{wagner2017case}.
The source code for the CS2SA* and MATLS solvers
was obtained from the corresponding authors. 

The relative performance of the solvers is shown in Table~\ref{tab:experiments_3}.
Overall, the results indicate that the techniques used by the CoCo solver are beneficial,
especially the proposed PGCH approach which provides explicit coordination for solving the interdependent KP and TSP components.
Using analysis of variance (ANOVA) on the corresponding overall gain values,
followed by t-tests with a confidence interval of 95\%~\cite{peck2019introduction},
the proposed CoCo solver obtained statistically significantly better results
than the next best method in the vast majority of cases.
Similarly, in each category,
the statistical significance of the differences between the average RDI values obtained for CoCo
and the next best solver were confirmed using a paired t-test with a confidence interval of 95\%.
Each statistically significant difference is marked with a star in Table~\ref{tab:experiments_3}.

\begin{table*}[!tb]
\centering
\caption
  {
  Performance comparison of the CoCo solver (Solver~4 in Table~\ref{tab:experiments_1}) against
  MATLS~\cite{mei2014improving},
  S5~\cite{faulkner2015approximate}
  and
  CS2SA*~\cite{el2018efficiently}
  solvers.
  Results are reported in terms of the relative deviation index (RDI)
  on 3 categories of TTP instances as per Table~\ref{tab:experiments_1}.
  Corresponding average overall gain values are shown in Tables~\ref{tab:exp3catA}, \ref{tab:exp3catB} and \ref{tab:exp3catC} in~\ref{sec:appendix}.
  Statistically significant differences between CoCo and the next best solver are marked with a star~({$^\star$}).
  }
\label{tab:experiments_3}
\vspace{-1.5ex}
\small
\begin{tabular}{|l|r|r|r|r||r|r|r|r||r|r|r|r|}
\hline
\multirow{2}{*}{\bf Instance} & \multicolumn{4}{c||}{\textbf{Category A}} & \multicolumn{4}{c||}{\textbf{Category B}} & \multicolumn{4}{c|}{\textbf{Category C}} \\  \cline{2-13}
                              & \multicolumn{1}{c|}{\rotatebox[origin=c]{0}{\textbf{\scriptsize MATLS}}} & \multicolumn{1}{c|}{\rotatebox[origin=c]{0}{\textbf{\scriptsize ~~~~~S5~~~~~}}} & \multicolumn{1}{c|}{\rotatebox[origin=c]{0}{\textbf{\scriptsize CS2SA*}}} & \multicolumn{1}{c||}{\rotatebox[origin=c]{0}{\textbf{\scriptsize CoCo}}} &   \multicolumn{1}{c|}{\rotatebox[origin=c]{0}{\textbf{\scriptsize MATLS}}} & \multicolumn{1}{c|}{\rotatebox[origin=c]{0}{\textbf{\scriptsize ~~~~~S5~~~~~}}} & \multicolumn{1}{c|}{\rotatebox[origin=c]{0}{\textbf{\scriptsize CS2SA*}}} & \multicolumn{1}{c||}{\rotatebox[origin=c]{0}{\textbf{\scriptsize CoCo}}} &  \multicolumn{1}{c|}{\rotatebox[origin=c]{0}{\textbf{\scriptsize MATLS}}} & \multicolumn{1}{c|}{\rotatebox[origin=c]{0}{\textbf{\scriptsize ~~~~~S5~~~~~}}} & \multicolumn{1}{c|}{\rotatebox[origin=c]{0}{\textbf{\scriptsize CS2SA*}}} & \multicolumn{1}{c|}{\rotatebox[origin=c]{0}{\textbf{\scriptsize CoCo}}} \\ \hline\hline

eil76 & 72.2 & \textbf{100} & 36.3 & \textbf{100} & \textbf{96.1} & 95.7 & 40.9 & 95.6 & \textbf{98.2} & 97.3 & 66.1&90.5 \\ 
kroA100 & 76.9 & 67.2 & 15.2 & {$^\star$}~\textbf{99.2} & 64.0 & 33.5 & 3.5 & {$^\star$}~\textbf{90.3} & 63.8 & 64.7 & 32.4& {$^\star$}~\textbf{88.0} \\
ch130 & 49.2 & 86.8 & 42.7 & {$^\star$}~\textbf{100} & 92.2 & 95.1 & 21.1 & {$^\star$}~\textbf{100} & 83.9 & \textbf{93.8} & 16.6& 92.9 \\
u159 & 61.5 & 80.0 & 49.4 & {$^\star$}~\textbf{90.8} & 85.8 & 96.7 & 66.5 & {$^\star$}~\textbf{98.9} & 33.9 & 70.1 & 24.3& {$^\star$}~\textbf{93.4} \\ \hline
a280 & 70.1 & 92.0 & 42.0 & {$^\star$}~\textbf{98.4} & 71.3 & 60.4 & 29.6 & {$^\star$}~\textbf{100} & 90.0 & 99.4 & 37.4&  {$^\star$}~\textbf{100} \\
u574 & 65.9 & 90.7 & 25.5 & {$^\star$}~\textbf{98.6} & 81.5 & 82.4 & 31.6 & {$^\star$}~\textbf{98.8} & 94.2 & 89.9 & 39.8& \textbf{95.2} \\
u724 & 41.5 & 76.0 & 16.4 & {$^\star$}~\textbf{97.1} & 52.8 & 60.3 & 39.1 & {$^\star$}~\textbf{95.5} & 68.5 & 73.8 & 26.4& {$^\star$}~\textbf{99.8} \\
dsj1000 & 92.4 & 4.8 & \textbf{100} & \textbf{100} & 43.6 & 53.4 & 16.8 & {$^\star$}~\textbf{98.0} & 61.3 & 90.3 & 45.3& {$^\star$}~\textbf{96.9} \\ \hline
rl1304 & 48.8 & 88.0 & 43.0 & {$^\star$}~\textbf{94.6} & 67.9 & 81.4 & 32.8 & {$^\star$}~\textbf{99.1} & 78.1 & 84.1 & 44.1& {$^\star$}~\textbf{98.7} \\
fl1577 & 54.6 & \textbf{93.4} & 15.6 & 92.9 & 75.8 & 84.2 & 50.4 & {$^\star$}~\textbf{93.3} & 88.1 & 89.4 & 43.0& {$^\star$}~\textbf{95.2} \\ 
d2103 & 1.0 & 85.7 & 63.3 & {$^\star$}~\textbf{92.7} & 32.3 & 68.0 & 27.0 & {$^\star$}~\textbf{98.3} & 25.2 & 48.5 & 21.6& {$^\star$}~\textbf{90.3} \\
pcb3038 & 41.4 & 91.4 & 19.4 & {$^\star$}~\textbf{97.4} & 46.2 & 63.3 & 32.9 & {$^\star$}~\textbf{96.3} & 77.7 & 85.7 & 45.6& {$^\star$}~\textbf{96.2} \\ \hline
fnl4461 & 30.6 & 84.4 & 5.3 & {$^\star$}~\textbf{91.5} & 82.1 & 86.8 & 56.4 & {$^\star$}~\textbf{97.8} & 93.2 & 91.7 & 27.3& {$^\star$}~\textbf{98.8} \\ 
pla7397 & 69.3 & 95.5 & 38.7 & {$^\star$}~\textbf{98.9} & 76.9 & 84.2 & 48.2 & {$^\star$}~\textbf{97.3} & 77.0 & 87.8 & 54.2& {$^\star$}~\textbf{96.7} \\
rl11849 & 27.3 & 87.0 & 8.7 & {$^\star$}~\textbf{95.1} & 54.7 & 62.1 & 32.6 & {$^\star$}~\textbf{93.5} & 53.1 & 52.2 & 22.5& {$^\star$}~\textbf{94.4} \\ 
usa13509 & 40.9 & 94.1 & 37.2 & {$^\star$}~\textbf{97.9} & 68.1 & 76.0 & 34.1 & {$^\star$}~\textbf{96.3} & 30.7 & 18.8 & 49.0& {$^\star$}~\textbf{83.8} \\ \hline
brd14051 & 41.5 & 91.4 & 19.7 & {$^\star$}~\textbf{96.4} & 69.9 & 75.1 & 49.4 & {$^\star$}~\textbf{96.0} & 59.6 & 72.0 & 56.0& {$^\star$}~\textbf{94.9} \\
d15112 & 2.9 & 75.7 & 21.6 & {$^\star$}~\textbf{95.8} & 19.4 & 43.9 & 52.3 & {$^\star$}~\textbf{85.9} & 11.0 & 22.2 & 48.8& {$^\star$}~\textbf{86.2} \\ 
d18512 & 51.5 & 93.9 & 20.6 & {$^\star$}~\textbf{97.5} & 81.4 & 86.5 & 57.0 & {$^\star$}~\textbf{97.9} & 57.3 & 60.6 & 43.7& {$^\star$}~\textbf{83.5} \\
pla33810 & 23.7 & 87.6 & 25.9 & {$^\star$}~\textbf{95.1} & 59.3 & 59.8 & 24.9 & {$^\star$}~\textbf{87.7} & 70.5 & 69.4 & 26.9& {$^\star$}~\textbf{96.5} \\ \hline\hline
Average & 48.2 & 83.3 & 32.3 & {$^\star$}~\textbf{96.5} & 66.1 & 72.4 & 37.4 & {$^\star$}~\textbf{95.8} & 65.8 & 73.1 & 38.5& {$^\star$}~\textbf{93.6} \\ \hline
\end{tabular}
\end{table*}

\clearpage

In the fourth set of experiments,
we compare the proposed CoCo solver against the recently proposed MEA2P solver~\cite{wuijts2019investigation},
which is a steady state memetic algorithm with edge-assembly~\cite{nagata2006new} and two-point crossover operators.
The MEA2P solver is the most recent solver targeted for solving small TTP instances such as~\cite{wagner2016stealing,mei2015heuristic,martins2017hseda,el2018hyperheuristic}.
The solver generates 50 solutions each with a random tour and an empty collection plan as the initial population.
Then, in each of the following 2500 iterations,
it combines two randomly selected solutions using the edge-assembly crossover operator~\cite{nagata2006new} on the tours
and the two-point crossover operator on the collection plans to generate a new solution.
The initial solutions and the following combined solutions are improved using a local search method via 2-OPT~\cite{croes1958method},
node insertion~\cite{faulkner2015approximate}, 
bit-flip~\cite{polyakovskiy2014comprehensive,faulkner2015approximate}
and item exchange~\cite{mei2016investigation} moves in an interleaved fashion
to address the interdependency between the TSP and KP components.

Due to the heavy computation demands of MEA2P and in order to allow a meaningful comparison,
instead of limiting the run time to 10 minutes,
we placed a limit on the number of restarts in the proposed CoCo solver to be 2500 restarts for solving each TTP instance.
Furthermore, we limit this experiment to the first eight small TTP instances from the three categories used in the  previous experiments,
due to the non-practicality of MEA2P for solving larger TTP instances
(ie.,~the MEA2P solver is too compuationally expensive for solving larger TTP instances in a reasonable amount of time).

The relative performance of the CoCo and MEA2P solvers is shown in Table~\ref{tab:experiments_4}.
The second column in the table shows the percentage of unique initial tours
generated via the Chained Lin-Kernighan heuristic~\cite{applegate2003chained} (in the proposed CoCo solver)
for solving the underlying TSP instance (being the same in all three categories).

The results indicate that in almost all the small instances
for which the Chained Lin-Kernighan heuristic generated a small number of unique initial tours,
the MEA2P solver performed better than the proposed CoCo solver.
We conjecture that using random initialisation with the edge-assembly and two-point crossover operators 
allows the MEA2P solver explore more diverse solutions in such small instances.
This is in contrast to the proposed CoCo solver, which relies on the Chained Lin-Kernighan heuristic
as well as Insertion~\cite{mei2014improving} and PackIterative~\cite{faulkner2015approximate} item collecting heuristics
to initialise the tour and the corresponding collection plan.
However, as the size of the problem increases, ie.,~an increase in the number of cities and items,
the Chained Lin-Kernighan heuristic generates sufficiently diverse initial tours,
and the MEA2P solver loses its advantage in exploring more diverse solutions.
The results indicate that despite MEA2P solver spending hours or even days to solve these larger instances,
it did not find better solutions than the CoCo solver which only required a few minutes.
 
\begin{table}[!tb]
\centering
\caption
  {
  Performance comparison of the CoCo solver (Solver~4 in Table~\ref{tab:experiments_1}) against the MEA2P~\cite{wuijts2019investigation} solver.
  The second column shows the percentage of unique initial tours generated by the Chained Lin-Kernighan heuristic (in our proposed CoCo solver).
  The first two columns in each category show the relative deviation index (RDI) values on 3 categories of 8 small TTP instances as per Table~\ref{tab:experiments_1}.
  Corresponding average overall gain values are shown in Table~\ref{tab:exp4} in~\ref{sec:appendix}.
  Statistically significant differences between CoCo and MEA2P are marked with a star~({$^\star$}).
  The last two columns in each category show the average time over 10 runs spent by each solver in seconds~(s), minutes~(m), hours~(h) or days~(d) to solve the corresponding instance. 
  }
\label{tab:experiments_4}
\vspace{-1.5ex}
\small
\setlength{\tabcolsep}{4.5pt}
\begin{tabular}{|l|r|r|r||r|r||r|r||r|r||r|r||r|r|}
\hline
\multirow{3}{*}{\bf Instance} & \textbf{\scriptsize Unique}~~&\multicolumn{4}{c||}{\textbf{Category A}} & \multicolumn{4}{c||}{\textbf{Category B}} & \multicolumn{4}{c|}{\textbf{Category C}} \\  \cline{3-14}

  & \textbf{\scriptsize Initial}~~ & \multicolumn{2}{c||}{\textbf{RDI Values}} & \multicolumn{2}{c||}{\textbf{Avg. Time}} &
      \multicolumn{2}{c||}{\textbf{RDI Values}} & \multicolumn{2}{c||}{\textbf{Avg. Time}} &
      \multicolumn{2}{c||}{\textbf{RDI Values}} & \multicolumn{2}{c|}{\textbf{Avg. Time}}
       \\ \cline{3-14}
 
& \textbf{\scriptsize Tours (\%)}&\multicolumn{1}{c|}{\rotatebox[origin=c]{0}{\textbf{\scriptsize MEA2P}}}  &
    \multicolumn{1}{c||}{\rotatebox[origin=c]{0}{\textbf{\scriptsize CoCo}}}  & 
    \multicolumn{1}{c|}{\rotatebox[origin=c]{0}{\textbf{\scriptsize MEA2P}}}  &
    \multicolumn{1}{c||}{\rotatebox[origin=c]{0}{\textbf{\scriptsize CoCo}}}  & 
    \multicolumn{1}{c|}{\rotatebox[origin=c]{0}{\textbf{\scriptsize MEA2P}}}  &
    \multicolumn{1}{c||}{\rotatebox[origin=c]{0}{\textbf{\scriptsize CoCo}}}  & 
    \multicolumn{1}{c|}{\rotatebox[origin=c]{0}{\textbf{\scriptsize MEA2P}}}  &
    \multicolumn{1}{c||}{\rotatebox[origin=c]{0}{\textbf{\scriptsize CoCo}}}  & 
    \multicolumn{1}{c|}{\rotatebox[origin=c]{0}{\textbf{\scriptsize MEA2P}}}  &
    \multicolumn{1}{c||}{\rotatebox[origin=c]{0}{\textbf{\scriptsize CoCo}}} & 
    \multicolumn{1}{c|}{\rotatebox[origin=c]{0}{\textbf{\scriptsize MEA2P}}} &
     \multicolumn{1}{c|}{\rotatebox[origin=c]{0}{\textbf{\scriptsize ~CoCo}}}  \\ \hline

eil76   &  3.3 & 41.4 & \textbf{54.6}  & 45s & 22.7s &  {$^\star$}~\textbf{95.7} & 0.0 & 2.3m & 28.3s & {$^\star$}~\textbf{100} & 23.6 &4.7m&40.6s\\ 
           
kroA100 &  1.2 & {$^\star$}~\textbf{94.8} & 5.6 & 93.6s & 41.9s &  {$^\star$}~\textbf{99.2} & 54.4 & 5m & 48.7s &  {$^\star$}~\textbf{100} & 0.0 & 10m & 1.1m \\
           
ch130   &  4.9 & {$^\star$}~\textbf{84.1} & 23.5 & 3m & 1.4m & {$^\star$}~\textbf{94.7} & 30.3 & 9.8m & 93.2s &  {$^\star$}~\textbf{79.7} & 8.4 & 20.3m & 2m   \\
           
u159    &  4.4 & {$^\star$}~\textbf{71.2} & 0.0 & 7.2m & 1m &  {$^\star$}~\textbf{82.2} & 12.4 & 16.8m & 76.4s &  {$^\star$}~\textbf{100} & 0.0 & 31.1m & 1.7m \\ \hline
           
a280    & 80.3 & {$^\star$}~\textbf{52.3} & 9.4 & 31.8m & 1.4m & 56.4 & {$^\star$}~\textbf{77.1} & 1.3h & 1.9m & 73.4 &  {$^\star$}~\textbf{100} & 2h & 2.6m \\ 

u574    & 65.1 & {$^\star$}~\textbf{61.1} & 8.9 & 4.5h & 9.1m &  {$^\star$}~\textbf{40.0} & 6.6 & 12.9h & 10.4m & 47.6 &  {$^\star$}~\textbf{74.0} & 25.2h & 11.8m \\

u724    & 90.2 & 31.8 &  {$^\star$}~\textbf{90.0} & 6.5h & 8.5m & 44.7 &  \textbf{50.4} & 28.3h & 11m & 48.5 & {$^\star$}~\textbf{76.0} & 2.1d & 13m   \\

dsj1000 & 87.3 & 9.8 &  {$^\star$}~\textbf{100} & 5.9h & 33.8m &  \textbf{40.9} & 34.3 & 3.4d & 38.2m & 62.3 &  {$^\star$}~\textbf{80.2} & 6.3d & 42.7m \\ \hline

\end{tabular}
\end{table}

Lastly, a comparison of the maximum {$G(c,p)$} values obtained by any variant of our proposed solver
against the best known maximum values reported in the literature~\cite{wuijts2019investigation,wagner2017case} 
(by enforcing a 10 minute limit for each instance)
is shown in Tables~\ref{tab:exp1catA}, \ref{tab:exp1catB} and \ref{tab:exp1catC} in~\ref{sec:appendix}.
In the vast majority of cases, our variants obtained higher {$G(c,p)$} values than previous solvers.

\clearpage
\section{Conclusion}
\label{sec:conclusion}

Many practical constraint optimisation problems
are comprised of several interdependent components.
Due to the interdependency,
simply finding an optimal solution to each underlying component
does not guarantee that the solution to the overall problem is optimal.
The travelling thief problem (TTP) is a representative of such multi-component problems.
Here, the thief performs a cyclic tour through a set of cities,
and pursuant to a collection plan,
collects a subset of obtainable items into a finite-capacity rented knapsack.
TTP can be thought of as a merger of two interdependent components:
the knapsack problem (KP) and the travelling salesman problem (TSP).
A TTP solution includes a cyclic tour through the cities as a solution to the TSP component,
and a plan for collecting items as a solution to the KP component.
Inspired by the co-operational co-evolution approach~\cite{potter1994cooperative}, 
methods to solving TTPs often involve solving the KP and TSP components in an interleaved manner
via dedicated component solvers~\cite{bonyadi2014socially}.
Solution to one component is kept fixed while the solution to the other component is modified.

In the TTP setting, items are scattered over the cities,
with the order of the cities in the tour restricting the collection order of the items.
As such, changing the order of the cities in the tour requires corresponding changes to the collection plan.
The 2-OPT segment reversal heuristic~\cite{croes1958method} is often employed for solving the TSP component.
As the length of the segment to be reversed increases,
the amount of corresponding changes required for the collection plan is likely to increase. 
Within the context of a meta-optimiser that interleaves solving the KP and TSP components,
changes to the collection plan are postponed until the dedicated KP solver is executed.
However, if a reversed segment is not accepted while solving the TSP component
(due to obtaining a lower objective value),
there is no opportunity to evaluate corresponding changes to the collection plan for the KP component.
This unnecessarily restricts the search space,
as potentially beneficial combinations of segment reversal with corresponding changes to the collection plan are not even attempted.

To address the above issue,
we have proposed a new heuristic for solving the TSP component,
termed Profit Guided Coordination Heuristic (PGCH).
When a segment in the cyclic tour is reversed, the collection plan is adjusted accordingly.
Items regarded as less profitable and collected in cities located earlier in the reversed segment
are substituted by items that tend to be equally or more profitable and not collected in cities located later in the reversed segment.
Using PGCH for solving the TSP component, segment reversing moves that are longer than \mbox{2-OPT} tend to be accepted.
As a result, the quality of the cyclic tour is considerably improved.

For solving the KP component, an often used approach is a hill-climber that searches via flipping the collection status
(from collected to uncollected and vice-versa)
of one item at a time~\cite{polyakovskiy2014comprehensive,faulkner2015approximate}.
Such a small and untargeted change in the collection plan results in a slow and meandering exploration of the solution space.
When a time limit is placed for finding a solution, the solution space may not be explored adequately,
which can contribute to low quality TTP solutions.
To make the search more targeted, we have proposed a modified form of bit-flip search (termed Boundary Bit-Flip)
where changes in the collection state are only permitted for boundary items.
Such items are defined as the lowest profitable collected items or highest profitable uncollected items.
This restriction reduces the amount of time spent on solving the KP component,
thereby allowing more tours to be evaluated by the TSP component within a given time budget.

The two proposed approaches (PGCH and Boundary Bit-Flip) form the basis of a new cooperative coordination (CoCo) solver.
On a comprehensive set of benchmark TTP instances,
the proposed CoCo solver has found new best gain values for most of the TTP instances
and has also outperformed several notable solvers:
MATLS~\cite{mei2014improving},
S5~\cite{faulkner2015approximate},
CS2SA*~\cite{el2018efficiently}
and
MEA2P~\cite{wuijts2019investigation}.
However, for some small instances for which the Chained Lin-Kernighan heuristic~\cite{applegate2003chained}
used by the CoCo solver does not generate sufficiently diverse initial tours,
MEA2P solver finds better solutions.
We are aiming to address this issue in future work.

\section*{Acknowledgements}

This article is a revised and extended version of our earlier work
published at the International Symposium on Combinatorial Search~\cite{namazi2019pgch}.
We would like to thank our colleagues at Data61/CSIRO (Toby Walsh, Phil Kilby, Regis Riveret)
for discussions leading to the improvement of this article.

\section*{Disclosure Statement}

Declarations of interest: none.









\bibliographystyle{elsarticle-num}

\clearpage
\bibliography{references}

\begin{thebibliography}{10}
\expandafter\ifx\csname url\endcsname\relax
  \def\url#1{\texttt{#1}}\fi
\expandafter\ifx\csname urlprefix\endcsname\relax\def\urlprefix{URL }\fi
\expandafter\ifx\csname href\endcsname\relax
  \def\href#1#2{#2} \def\path#1{#1}\fi

\bibitem{namazi2019pgch}
M.~Namazi, M.~A.~H. Newton, A.~Sattar, C.~Sanderson, A profit guided
  coordination heuristic for travelling thief problems, in: International
  Symposium on Combinatorial Search, AAAI, 2019, pp. 140--144.

\bibitem{rossi2006handbook}
F.~Rossi, P.~Van~Beek, T.~Walsh, Handbook of Constraint Programming, Elsevier,
  2006.

\bibitem{bonyadi2013travelling}
M.~R. Bonyadi, Z.~Michalewicz, L.~Barone, The travelling thief problem: The
  first step in the transition from theoretical problems to realistic problems,
  in: IEEE Congress on Evolutionary Computation (CEC), 2013, pp. 1037--1044.

\bibitem{michalewicz2012quo}
Z.~Michalewicz, Quo vadis, evolutionary computation?, in: Advances in
  Computational Intelligence, Lecture Notes in Computer Science (LNCS), Vol.
  7311, 2012, pp. 98--121.

\bibitem{mei2016investigation}
Y.~Mei, X.~Li, X.~Yao, On investigation of interdependence between sub-problems
  of the travelling thief problem, Soft Computing 20~(1) (2016) 157--172.

\bibitem{bonyadi2019evolutionary}
M.~R. Bonyadi, Z.~Michalewicz, M.~Wagner, F.~Neumann, Evolutionary computation
  for multicomponent problems: opportunities and future directions, in:
  Optimization in Industry, Springer, 2019, pp. 13--30.

\bibitem{polyakovskiy2014comprehensive}
S.~Polyakovskiy, M.~R. Bonyadi, M.~Wagner, Z.~Michalewicz, F.~Neumann, A
  comprehensive benchmark set and heuristics for the traveling thief problem,
  in: Annual Conference on Genetic and Evolutionary Computation, 2014, pp.
  477--484.

\bibitem{inbookKP}
H.~Kellerer, U.~Pferschy, D.~Pisinger, Introduction to {NP}-completeness of
  knapsack problems, in: Knapsack Problems, Springer, 2004, pp. 483--493.

\bibitem{gutin2006traveling}
G.~Gutin, A.~P. Punnen, The Traveling Salesman Problem and Its Variations,
  Springer, 2006.

\bibitem{mei2014improving}
Y.~Mei, X.~Li, X.~Yao, Improving efficiency of heuristics for the large scale
  traveling thief problem, in: Simulated Evolution and Learning, Lecture Notes
  in Computer Science (LNCS), Vol. 8886, 2014, pp. 631--643.

\bibitem{croes1958method}
G.~A. Croes, A method for solving traveling-salesman problems, Operations
  Research 6~(6) (1958) 791--812.

\bibitem{faulkner2015approximate}
H.~Faulkner, S.~Polyakovskiy, T.~Schultz, M.~Wagner, Approximate approaches to
  the traveling thief problem, in: Annual Conference on Genetic and
  Evolutionary Computation, 2015, pp. 385--392.

\bibitem{el2018efficiently}
M.~El~Yafrani, B.~Ahiod, Efficiently solving the {T}raveling {T}hief {P}roblem
  using hill climbing and simulated annealing, Information Sciences 432 (2018)
  231--244.

\bibitem{wuijts2019investigation}
R.~H. Wuijts, D.~Thierens, Investigation of the traveling thief problem, in:
  Proceedings of the Genetic and Evolutionary Computation Conference, 2019, pp.
  329--337.

\bibitem{wagner2017case}
M.~Wagner, M.~Lindauer, M.~M{\i}s{\i}r, S.~Nallaperuma, F.~Hutter, A case study
  of algorithm selection for the traveling thief problem, Journal of Heuristics
  24~(3) (2018) 295--320.

\bibitem{applegate2003chained}
D.~Applegate, W.~Cook, A.~Rohe, Chained {L}in-{K}ernighan for large traveling
  salesman problems, INFORMS Journal on Computing 15~(1) (2003) 82--92.

\bibitem{bonyadi2014socially}
M.~R. Bonyadi, Z.~Michalewicz, M.~R. Przybylek, A.~Wierzbicki, Socially
  inspired algorithms for the travelling thief problem, in: Annual Conference
  on Genetic and Evolutionary Computation, 2014, pp. 421--428.

\bibitem{potter1994cooperative}
M.~A. Potter, K.~A. De~Jong, A cooperative coevolutionary approach to function
  optimization, in: International Conference on Parallel Problem Solving from
  Nature, Springer, 1994, pp. 249--257.

\bibitem{el2016population}
M.~El~Yafrani, B.~Ahiod, Population-based vs. single-solution heuristics for
  the travelling thief problem, in: Proceedings of the Genetic and Evolutionary
  Computation Conference, ACM, 2016, pp. 317--324.

\bibitem{wagner2016stealing}
M.~Wagner, Stealing items more efficiently with ants: a swarm intelligence
  approach to the travelling thief problem, in: Swarm Intelligence, Lecture
  Notes in Computer Science (LNCS), Vol. 9882, 2016, pp. 273--281.

\bibitem{stutzle2000max}
T.~St{\"u}tzle, H.~H. Hoos, {MAX-MIN} ant system, Future Generation Computer
  Systems 16~(8) (2000) 889--914.

\bibitem{el2017local}
M.~El~Yafrani, B.~Ahiod, A local search based approach for solving the
  {T}ravelling {T}hief {P}roblem: The pros and cons, Applied Soft Computing 52
  (2017) 795--804.

\bibitem{mei2015heuristic}
Y.~Mei, X.~Li, F.~Salim, X.~Yao, Heuristic evolution with genetic programming
  for traveling thief problem, in: IEEE Congress on Evolutionary Computation
  (CEC), 2015, pp. 2753--2760.

\bibitem{martins2017hseda}
M.~S. Martins, M.~El~Yafrani, M.~R. Delgado, M.~Wagner, B.~Ahiod,
  R.~L{\"u}ders, {HSEDA}: A heuristic selection approach based on estimation of
  distribution algorithm for the travelling thief problem, in: Proceedings of
  the Genetic and Evolutionary Computation Conference, ACM, 2017, pp. 361--368.

\bibitem{el2018hyperheuristic}
M.~El~Yafrani, M.~Martins, M.~Wagner, B.~Ahiod, M.~Delgado, R.~L{\"u}ders, A
  hyperheuristic approach based on low-level heuristics for the travelling
  thief problem, Genetic Programming and Evolvable Machines 19~(1-2) (2018)
  121--150.

\bibitem{delaunay1934sphare}
B.~Delaunay, Sur la sph\`{e}re vide, Izvestia Akademii Nauk {SSSR}, Otdelenie
  Matematicheskikh i Estestvennykh Nauk 7 (1934) 793--800.

\bibitem{reinelt1991tsplib}
G.~Reinelt, {TSPLIB-A} traveling salesman problem library, {ORSA} Journal on
  Computing 3~(4) (1991) 376--384.

\bibitem{kim1996simulated}
J.-U. Kim, Y.-D. Kim, Simulated annealing and genetic algorithms for scheduling
  products with multi-level product structure, Computers \& Operations Research
  23~(9) (1996) 857--868.

\bibitem{peck2019introduction}
R.~Peck, T.~Short, C.~Olsen, Introduction to statistics and data analysis,
  Cengage Learning, 2019.

\bibitem{nagata2006new}
Y.~Nagata, New {EAX} crossover for large {TSP} instances, in: Parallel Problem
  Solving from Nature - {PPSN~IX}, Springer, 2006, pp. 372--381.

\end{thebibliography}

\newpage
\appendix
\section{Overall Gain Values}
\label{sec:appendix}

\setcounter{table}{0}
\renewcommand{\thetable}{A\arabic{table}}

Tables~\ref{tab:exp1catA} through to~\ref{tab:exp4} 
provide the overall gain values
used for obtaining the RDI values in Section~\ref{sec:experiments}.
Furthermore, the last column in Tables~\ref{tab:exp1catA},~\ref{tab:exp1catB} and~\ref{tab:exp1catC}
shows the best gain values found by any solver 
in the following set:
the MEA2P solver~\cite{wuijts2019investigation} and the 21 solvers as reported in~\cite{wagner2017case}
(excluding the CS2SA solver~\cite{el2016population} due to a faulty evaluation in~\cite{wagner2017case}, as reported in~\cite{wuijts2019investigation}).
Best gain values are marked with a bullet ($\bullet$).
\begin{table}[htbp]
\caption
  {
  Mean overall gain values each over 10 runs corresponding to the solvers in Table \ref{tab:experiments_1}
  with the minimum and maximum overall gain values among all 40 runs on Category A instances.
  The last column shows the best gain values found by any solver in the following set:
  the MEA2P solver~\cite{wuijts2019investigation} and the 21 solvers as reported in~\cite{wagner2017case}
  (excluding the CS2SA solver~\cite{el2016population} due to a faulty evaluation in~\cite{wagner2017case}, as reported in~\cite{wuijts2019investigation}).
  Best gain values are marked with a bullet ($\bullet$).
  }
\centering
\footnotesize
\begin{tabular}{|l|r|r|r|r|r|r||r|}
\hline
\multirow{2}{*}{\bf Instance}&\multicolumn{6}{c||}{\textbf{Category A}}&\multicolumn{1}{c|}{\textbf{Best Gain by}}\\ \cline{2-7}
 & \multicolumn{1}{r|}{\textbf{Solver 1}} & \multicolumn{1}{r|}{\textbf{Solver 2}} & \multicolumn{1}{r|}{\textbf{Solver 3}} & \multicolumn{1}{r|}{\textbf{Solver 4}} & \multicolumn{1}{r|}{\textbf{Min}}&\multicolumn{1}{r||}{\textbf{Max}}& \multicolumn{1}{r|}{\textbf{Other Solvers}}\\ \hline\hline
eil76 & \textbf{4109} & \textbf{4109} & \textbf{4109} & \textbf{4109} & 4109 & $\bullet$~{4109}& $\bullet$~{4109} \\
kroA100 & 4740 & 4808 & \textbf{4855} & 4827 & 4601&4868& $\bullet$~{4976} \\ 
ch130 & 9520 & 9506 & \textbf{9564} & \textbf{9564} & 9470&9564& $\bullet$~{9707} \\ 
u159 & 8635 & 8634 & \textbf{8842} & 8820 & 8634&8979&$\bullet$~{9064} \\ \hline
a280 & 18441 & 18437 & \textbf{18668} & 18636 & 18437& $\bullet$~{18741}& 18441 \\ 
u574 & 27243 & 27140 & 27360 & \textbf{27367} & 26987& $\bullet$~{27517}& 27238 \\ 
u724 & 50353 & 50425 & \textbf{51149} & 51124 & 50274& $\bullet$~{51522}& 50346 \\ 
dsj1000 & \textbf{144219} & \textbf{144219} & \textbf{144219} & \textbf{144219} & 144219& $\bullet$~{144219}& 144118 \\ \hline
rl1304 & 81283 & 81353 & 81611 & \textbf{81769} & 80862& $\bullet$~{82470}& 80050 \\ 
fl1577 & 93135 & 93006 & \textbf{93392} & 93030 & 91702& $\bullet$~{93934}& 90300 \\ 
d2103 & 121533 & 121411 & \textbf{122175} & 121977 & 120874& $\bullet$~{122719}& 120675 \\
pcb3038 & 160566 & 160501 & \textbf{161805} & 161456 & 160174& $\bullet$~{162113}& 160196 \\ \hline
fnl4461 & 262813 & 263017 & \textbf{264334} & 264332 & 261929& $\bullet$~{266549}& 263040 \\ 
pla7397 & 397712 & 397800 & 399223 & \textbf{399349} & 393387& $\bullet$~{401448}&392980  \\
rl11849 & 709345 & 709222 & 712867 & \textbf{714396} & 707582& $\bullet$~{718089}& 706973 \\
usa13509 & 809231 & 809420 & \textbf{813417} & 813324 & 805792& $\bullet$~{815853}& 804847 \\ \hline
brd14051 & 875029 & 877006 & \textbf{883158} & 881904 & 872146& $\bullet$~{888011}& 874656 \\ 
d15112 & 950711 & 949399 & \textbf{968331} & 967701 & 943920& $\bullet$~{972050}& 938615 \\ 
d18512 & 1075706 & 1072948 & \textbf{1082126} & 1079883 &1069638& $\bullet$~{1086072}&1071013 \\
pla33810 & 1899060 & 1897881 & 1913638 & \textbf{1917177} & 1882973& $\bullet$~{1929990}& 1863668 \\ \hline
\end{tabular}
\label{tab:exp1catA}
\end{table}

\begin{table}[htbp]
\caption
  {
  Mean overall gain values each over 10 runs corresponding to the solvers in Table \ref{tab:experiments_1}
  with the minimum and maximum overall gain values among all 40 runs on Category B instances.
  The last column shows the best gain values found by any solver in the following set:
  the MEA2P solver~\cite{wuijts2019investigation} and the 21 solvers as reported in~\cite{wagner2017case}
  (excluding the CS2SA solver~\cite{el2016population} due to a faulty evaluation in~\cite{wagner2017case}, as reported in~\cite{wuijts2019investigation}).
  Best gain values are marked with a bullet ($\bullet$).
  }
\centering
\footnotesize
\begin{tabular}{|l|r|r|r|r|r|r||r|}
\hline
\multirow{2}{*}{\bf Instance}&\multicolumn{6}{c||}{\textbf{Category B}}&\multicolumn{1}{c|}{\textbf{Best Gain by}}\\ \cline{2-7}
 & \multicolumn{1}{r|}{\textbf{Solver 1}} & \multicolumn{1}{r|}{\textbf{Solver 2}} & \multicolumn{1}{r|}{\textbf{Solver 3}} & \multicolumn{1}{r|}{\textbf{Solver 4}} & \multicolumn{1}{r|}{\textbf{Min}}&\multicolumn{1}{r||}{\textbf{Max}}& \multicolumn{1}{r|}{\textbf{Other Solvers}}\\ \hline\hline
eil76 & 21647 & 21349 & \textbf{22269} & 22240 & 20854&22318& $\bullet$~{23278}\\ 
kroA100 & 41330 & 41365 & \textbf{45503} & 45139 & 41241&45812&$\bullet$~{46633}\\ 
ch130 & 61290 & 61290 & 61623 & \textbf{61702} & 61286&61703&$\bullet$~{62496}\\ 
u159 & 60719 & \textbf{61016} & 60770 & 60897 & 60626&$\bullet$~{61077}& 60630 \\ \hline
a280 & 110180 & 110302 & 116454 & \textbf{116457} & 110151&$\bullet$~{116458}& 110147 \\ 
u574 & 256850 & 257399 & \textbf{260198} & 260062 & 256245&$\bullet$~{261036}& 252685 \\ 
u724 & 307831 & 308087 & 321335 & \textbf{322133} & 306650&$\bullet$~{324316}& 305029 \\ 
dsj1000 & 347065 & 347201 & 370065 & \textbf{370435} & 345482&$\bullet$~{373254}& 352185 \\ \hline
rl1304 & 577423 & 585897 & 598058 & \textbf{598610} & 577335&$\bullet$~{599351}& 573672 \\ 
fl1577 & 615705 & 619338 & 630214 & \textbf{635696} & 599956&$\bullet$~{648171}& 603331 \\ 
d2103 & 892467 & 896308 & 923100 & \textbf{925808} & 884237&$\bullet$~{927958}& 859121 \\ 
pcb3038 & 1180523 & 1182342 & 1198340 & \textbf{1204897} & 1173589&$\bullet$~{1207834}& 1180985 \\ \hline
fnl4461 & 1624169 & 1629758 & 1640806 & \textbf{1646520} & 1617244&$\bullet$~{1650698}& 1624298 \\ 
pla7397 & 4303587 & 4394070 & 4424384 & \textbf{4470928} & 4115015&$\bullet$~{4499883}& 4400966 \\ 
rl11849 & 4576428 & 4633472 & 4723718 & \textbf{4780661} & 4528579&$\bullet$~{4813852}& 4645464 \\ 
usa13509 & 7801763 & 8161340 & 8118937 & \textbf{8254572} & 6848356&$\bullet$~{8340377}& 7844821 \\ \hline
brd14051 & 6428360 & 6615466 & 6740590 & \textbf{6816674} & 6090832&$\bullet$~{6870081}& 6536560 \\ 
d15112 & 7051668 & 7393476 & 7474091 & \textbf{7714270} & 6576653&$\bullet$~{7909260}& 7026975 \\ 
d18512 & 6847516 & 7247641 & 7391191 & \textbf{7467890} & 6269585&$\bullet$~{7523963}& 7281727 \\ 
pla33810 & 15153466 & 15460892 & 16010375 & \textbf{16172664} & 14983123&$\bullet$~{16418169}& 15634853 \\ \hline
\end{tabular}
\label{tab:exp1catB}
\end{table}

\begin{table}[htbp]
\caption
  {
  Mean overall gain values each over 10 runs corresponding to the solvers in Table \ref{tab:experiments_1}
  with the minimum and maximum overall gain values among all 40 runs on Category C instances.
  The last column shows the best gain values found by any solver in the following set:
  the MEA2P solver~\cite{wuijts2019investigation} and the 21 solvers as reported in~\cite{wagner2017case}
  (excluding the CS2SA solver~\cite{el2016population} due to a faulty evaluation in~\cite{wagner2017case}, as reported in~\cite{wuijts2019investigation}).
  Best gain values are marked with a bullet ($\bullet$).
  }
\centering
\footnotesize
\begin{tabular}{|l|r|r|r|r|r|r||r|}
\hline
\multirow{2}{*}{\bf Instance}&\multicolumn{6}{c||}{\textbf{Category C}}&\multicolumn{1}{c|}{\textbf{Best Gain by}}\\ \cline{2-7}
 & \multicolumn{1}{r|}{\textbf{Solver 1}} & \multicolumn{1}{r|}{\textbf{Solver 2}} & \multicolumn{1}{r|}{\textbf{Solver 3}} & \multicolumn{1}{r|}{\textbf{Solver 4}} & \multicolumn{1}{r|}{\textbf{Min}}&\multicolumn{1}{r||}{\textbf{Max}}& \multicolumn{1}{r|}{\textbf{Other Solvers}}\\ \hline\hline
eil76 & \textbf{87476} & 87323 & 87270 & 87249 & 86366&88062& $\bullet$~{ 88386}\\ 
kroA100 & 155977 & 155621 & \textbf{157735} & 157712 & 155585&158812& $\bullet$~{159135}\\ 
ch130 & 207159 & 207081 & \textbf{207530} & 206851 & 206381&$\bullet$~{207902} & 207654 \\ 
u159 & 246602 & 246493 & \textbf{248815} & 248627 & 246132&$\bullet$~{249312} & 246572 \\ \hline
a280 & 429095 & 429095 & 429136 & \textbf{429138} & 429082&$\bullet$~{429138} & 429082 \\ 
u574 & 967068 & 967355 & \textbf{969973} & 969666 & 966227&$\bullet$~{970692} & 966853 \\ 
u724 & 1192048 & 1191552 & 1203543 & \textbf{1204366} & 1187571&$\bullet$~{1205747} & 1188503 \\ 
dsj1000 & 1484785 & 1487642 & 1490023 & \textbf{1495584} & 1480693 & $\bullet$~{1500167} & 1477691 \\ \hline
rl1304 & 2188104 & 2188936 & 2209944 & \textbf{2212260} & 2183617&$\bullet$~{2214083} & 2183617 \\ 
fl1577 & 2462610 & 2479097 & 2482379 & \textbf{2491120} & 2444829&$\bullet$~{2504641} & 2469869 \\ 
d2103 & 3454687 & 3457755 & 3489555 & \textbf{3493856} & 3447868&$\bullet$~{3507989} & 3385940 \\ 
pcb3038 & 4564948 & 4572248 & 4591221 & \textbf{4596155} & 4553969&$\bullet$~{4607306} & 4578247 \\ \hline
fnl4461 & 6539711 & 6553039 & 6563496 & \textbf{6569715} & 6526013& $\bullet$~{6573564} & 6555400 \\ 
pla7397 & 13873075 & 14393277 & 14258323 & \textbf{14496916} & 13328394&$\bullet$~{14600106} &14282055 \\ 
rl11849 & 18196554 & 18268196 & 18398981 & \textbf{18512214} & 18093864&$\bullet$~{18544745} &18283865  \\
usa13509 & 25496366 & 26391280 & 26122134 & \textbf{26572209} & 24328292&$\bullet$~{26743719} & 26050289 \\\hline
brd14051 & 23065915 & 23869490 & 24082233 & \textbf{24188367} & 21942204& $\bullet$~{24260278} & 23846276 \\
d15112 & 26005329 & 27232945 & 27146384 & \textbf{27437349} & 25246146& $\bullet$~{27695658} &26168093  \\
d18512 & 25478702 & 27373733 & 27433958 & \textbf{27690630} & 24056519& $\bullet$~{27946341} &27429093 \\
pla33810 & 56305249 & 57695280 & 58151982 & \textbf{58744246} & 55320897& $\bullet$~{58857245} & 58236645 \\ \hline
\end{tabular}
\label{tab:exp1catC}
\end{table}

\begin{table}[htbp]
\caption
  {
  Mean overall gain values each over 10 runs corresponding to the solvers in Table \ref{tab:experiments_3}
  with the minimum and maximum overall gain values among all 40 runs on Category A instances.
  }
\centering
\footnotesize
\begin{tabular}{|l|r|r|r|r|r|r|}
\hline
\multirow{2}{*}{\bf Instance}&\multicolumn{6}{c|}{\textbf{Category A}}\\ \cline{2-7}
 & \multicolumn{1}{r|}{\textbf{MATLS}} & \multicolumn{1}{r|}{\textbf{S5}} & \multicolumn{1}{r|}{\textbf{CS2SA*}} & \multicolumn{1}{r|}{\textbf{CoCo}} & \multicolumn{1}{r|}{\textbf{Min}}&\multicolumn{1}{r|}{\textbf{Max}} \\ \hline\hline
eil76 & 3717 & \textbf{4109} & 3209 & \textbf{4109} & 2697&4109 \\ 
kroA100 & 4703 & 4650 & 4362 & \textbf{4827} & 4278&4831 \\ 
ch130 & 8868 & 9382 & 8779 & \textbf{9564} & 8193&9564 \\ 
u159 & 8314 & 8634 & 8105 & \textbf{8820} & 7252&8979 \\ \hline
a280 & 17639 & 18411 & 16648 & \textbf{18636} & 15169&18692 \\ 
u574 & 25881 & 27007 & 24043 & \textbf{27367} & 22886&27430 \\ 
u724 & 48865 & 50265 & 47839 & \textbf{51124} & 47174&51244 \\ 
dsj1000 & 143699 & 137740 & \textbf{144219} & \textbf{144219} & 137410&144219 \\ \hline
rl1304 & 75804 & 80911 & 75040 & \textbf{81769} & 69445&82470 \\ 
fl1577 & 88330 & \textbf{93081} & 83555 & 93030 & 81638&93895 \\ 
d2103 & 112686 & 121274 & 118997 & \textbf{121977} & 112584&122719 \\ 
pcb3038 & 148988 & 160115 & 144099 & \textbf{161456} & 139791&162028 \\ \hline
fnl4461 & 248482 & 262478 & 241881 & \textbf{264332} & 240500&266549 \\ 
pla7397 & 367247 & 395655 & 333910 & \textbf{399349} & 291891&400596 \\ 
rl11849 & 662940 & 708215 & 648843 & \textbf{714396} & 642243&718089 \\ 
usa13509 & 743146 & 808583 & 738638 & \textbf{813324} & 692847&815853 \\ \hline
brd14051 & 813977 & 875741 & 786919 & \textbf{881904} & 762545&886332 \\ 
d15112 & 871348 & 946896 & 890746 & \textbf{967701} & 868378&972050 \\ 
d18512 & 996820 & 1073310 & 941041 & \textbf{1079883} & 903748&1084385 \\ 
pla33810 & 1730997 & 1897560 & 1736746 & \textbf{1917177} & 1669331&1929990 \\ \hline
\end{tabular}
\label{tab:exp3catA}
\end{table}

\begin{table}[htbp]
\caption
  {
  Mean overall gain values each over 10 runs corresponding to the solvers in Table \ref{tab:experiments_3}
  with the minimum and maximum overall gain values among all 40 runs on Category B instances.
  }
\centering
\footnotesize
\begin{tabular}{|l|r|r|r|r|r|r|}
\hline
\multirow{2}{*}{\bf Instance}&\multicolumn{6}{c|}{\textbf{Category B}}\\ \cline{2-7}
 & \multicolumn{1}{r|}{\textbf{MATLS}} & \multicolumn{1}{r|}{\textbf{S5}} & \multicolumn{1}{r|}{\textbf{CS2SA*}} & \multicolumn{1}{r|}{\textbf{CoCo}} & \multicolumn{1}{r|}{\textbf{Min}}&\multicolumn{1}{r|}{\textbf{Max}} \\ \hline\hline
eil76 & \textbf{22286} & 22255 & 17484 & 22240 & 13932&22626 \\ 
kroA100 & 43310 & 41192 & 39109 & \textbf{45139} & 38865&45812 \\ 
ch130 & 60705 & 61071 & 51572 & \textbf{61702} & 48856&61703 \\ 
u159 & 58778 & 60550 & 55662 & \textbf{60897} & 44919&61077 \\ \hline
a280 & 111728 & 109932 & 104863 & \textbf{116457} & 99991&116458 \\ 
u574 & 252294 & 252698 & 229874 & \textbf{260062} & 215702&260621 \\ 
u724 & 303046 & 306396 & 296928 & \textbf{322133} & 279424&324139 \\
dsj1000 & 339630 & 345179 & 324412 & \textbf{370435} & 314896&371589 \\ \hline
rl1304 & 573980 & 584641 & 546284 & \textbf{598610} & 520355&599351 \\ 
fl1577 & 603443 & 618933 & 556335 & \textbf{635696} & 463057&648171 \\ 
d2103 & 842566 & 887644 & 835970 & \textbf{925808} & 801902&927958 \\ 
pcb3038 & 1164547 & 1178341 & 1153843 & \textbf{1204897} & 1127401&1207834 \\ \hline
fnl4461 & 1616072 & 1625086 & 1566075 & \textbf{1646520} & 1456772&1650698 \\ 
pla7397 & 4248266 & 4327733 & 3937097 & \textbf{4470928} & 3412869&4499883 \\ 
rl11849 & 4583308 & 4620657 & 4470733 & \textbf{4780661} & 4304699&4813852 \\ 
usa13509 & 7770907 & 7907496 & 7189622 & \textbf{8254572} & 6604595&8317858 \\ \hline
brd14051 & 6467038 & 6535639 & 6191198 & \textbf{6816674} & 5528880&6870081 \\ 
d15112 & 6797757 & 7135697 & 7251647 & \textbf{7714270} & 6530450&7909260 \\ 
d18512 & 7102985 & 7215763 & 6562282 & \textbf{7467890} & 5296681&7515204 \\
pla33810 & 15605360 & 15615480 & 14918919 & \textbf{16172664} & 14420806&16418169 \\ \hline
\end{tabular}
\label{tab:exp3catB}
\end{table}

\begin{table}[htbp]
\caption
  {
  Mean overall gain values each over 10 runs corresponding to the solvers in Table \ref{tab:experiments_3}
  with the minimum and maximum overall gain values among all 40 runs on Category C instances.
  }
\centering
\footnotesize
\begin{tabular}{|l|r|r|r|r|r|r|}
\hline
\multirow{2}{*}{\bf Instance}&\multicolumn{6}{c|}{\textbf{Category C}}\\ \cline{2-7}
 & \multicolumn{1}{r|}{\textbf{MATLS}} & \multicolumn{1}{r|}{\textbf{S5}} & \multicolumn{1}{r|}{\textbf{CS2SA*}} & \multicolumn{1}{r|}{\textbf{CoCo}} & \multicolumn{1}{r|}{\textbf{Min}}&\multicolumn{1}{r|}{\textbf{Max}} \\ \hline\hline
eil76 & \textbf{88131} & 88025 & 84476 & 87249 & 76964&88332 \\ 
kroA100 & 155500 & 155582 & 152621 & \textbf{157712} & 149656&158812 \\ 
ch130 & 205552 & \textbf{206981} & 195823 & 206851 & 193430&207881 \\ 
u159 & 242452 & 246212 & 241458 & \textbf{248627} & 238934&249312 \\ \hline
a280 & 426951 & 429014 & 415418 & \textbf{429138} & 407221&429138 \\ 
u574 & 969064 & 966602 & 937495 & \textbf{969666} & 914425&972457 \\ 
u724 & 1187092 & 1190001 & 1163870 & \textbf{1204366} & 1149297&1204466 \\
dsj1000 & 1442153 & 1485611 & 1418126 & \textbf{1495584} & 1350190&1500167 \\ \hline
rl1304 & 2183137 & 2191679 & 2135303 & \textbf{2212260} & 2073066&2214083 \\ 
fl1577 & 2470833 & 2474542 & 2342494 & \textbf{2491120} & 2220287&2504641 \\ 
d2103 & 3399342 & 3433231 & 3394085 & \textbf{3493856} & 3362770&3507989 \\ 
pcb3038 & 4553407 & 4571841 & 4479238 & \textbf{4596155} & 4373958&4604839 \\ \hline
fnl4461 & 6552392 & 6547803 & 6348407 & \textbf{6569715} & 6263913&6573564 \\ 
pla7397 & 13891830 & 14223020 & 13187494 & \textbf{14496916} & 11517832&14600106 \\ 
rl11849 & 18273340 & 18267920 & 18096006 & \textbf{18512214} & 17966059&18544745 \\ 
usa13509 & 26010780 & 25884060 & 26203582 & \textbf{26572209} & 25685658&26743719 \\ \hline
brd14051 & 23687540 & 23863340 & 23635912 & \textbf{24188367} & 22841916&24260278 \\ 
d15112 & 26032000 & 26242890 & 26738248 & \textbf{27437349} & 25827400&27695658 \\ 
d18512 & 27282770 & 27333880 & 27072529 & \textbf{27690630} & 26393417&27946341 \\ 
pla33810 & 57896250 & 57861950 & 56476214 & \textbf{58744246} & 55601509&58857245 \\ \hline
\end{tabular}
\label{tab:exp3catC}
\end{table}

\begin{table}[htbp]
\caption
  {
  Mean overall gain values each over 10 runs corresponding to the solvers in Table \ref{tab:experiments_4}
  with the minimum and maximum overall gain values among all 20 runs on 8 small instances in all there categories.
  }
\label{tab:exp4}
\vspace{-1.5ex}
\small
\setlength{\tabcolsep}{3.8pt}
\begin{tabular}{|l|r|r||r|r||r|r||r|r||r|r||r|r|}
\hline
\multirow{3}{*}{\bf Instance} & \multicolumn{4}{c||}{\textbf{Category A}} & \multicolumn{4}{c||}{\textbf{Category B}} & \multicolumn{4}{c|}{\textbf{Category C}} \\  \cline{2-13}

&  \multicolumn{1}{c|}{\rotatebox[origin=c]{0}{\textbf{\scriptsize MEA2P}}}  &
    \multicolumn{1}{c||}{\rotatebox[origin=c]{0}{\textbf{\scriptsize CoCo}}}  & 
    \multicolumn{1}{c|}{\rotatebox[origin=c]{0}{\textbf{\scriptsize Min}}}  &
    \multicolumn{1}{c||}{\rotatebox[origin=c]{0}{\textbf{\scriptsize Max}}}  & 
    \multicolumn{1}{c|}{\rotatebox[origin=c]{0}{\textbf{\scriptsize MEA2P}}}  &
    \multicolumn{1}{c||}{\rotatebox[origin=c]{0}{\textbf{\scriptsize CoCo}}}  & 
    \multicolumn{1}{c|}{\rotatebox[origin=c]{0}{\textbf{\scriptsize Min}}}  &
    \multicolumn{1}{c||}{\rotatebox[origin=c]{0}{\textbf{\scriptsize Max}}}  & 
    \multicolumn{1}{c|}{\rotatebox[origin=c]{0}{\textbf{\scriptsize MEA2P}}}  &
    \multicolumn{1}{c||}{\rotatebox[origin=c]{0}{\textbf{\scriptsize CoCo}}} & 
    \multicolumn{1}{c|}{\rotatebox[origin=c]{0}{\textbf{\scriptsize Min}}} &
     \multicolumn{1}{c|}{\rotatebox[origin=c]{0}{\textbf{\scriptsize ~Max}}}  \\ \hline

eil76	&4017	&\textbf{4038}	&3952	&4109	&\textbf{23228}	&22114	&22114	&23278	&\textbf{88386}	&86843	&86366	&88386\\
kroA100	&\textbf{4956}	&4626	&4605	&4976	&\textbf{46605}	&44970	&42983	&46633	&\textbf{159135}	&155585	&155585	&159135\\
ch130	&\textbf{9635}	&9456	&9386	&9682	&\textbf{62403}	&61283	&60756	&62496	&\textbf{207569}	&206383	&206242	&207907\\
u159	&\textbf{8977}	&8763	&8763	&9064	&\textbf{61820}	&60495	&60261	&62157	&\textbf{252667}	&248133	&248133	&252667\\ \hline
a280	&\textbf{18961}	&18547	&18456	&19422	&116084	&\textbf{116429}	&115145	&116810	&427369	&\textbf{429137}	&422492	&429138\\
u574	&\textbf{28560}	&27358	&27152	&29457	&\textbf{263011}	&260392	&259874	&267719	&964333	&\textbf{970229}	&953724	&976021\\
u724	&49866	&\textbf{51329}	&49067	&51579	&320703	&\textbf{321263}	&316293	&326151	&1197832 &\textbf{1204973} &1185219 &1211213\\
dsj1000	&134750	&\textbf{144219}	&133724	&144219	&\textbf{372407}	&371176	&364690	&383577	&1493313 &\textbf{1500002} &1470097 &1507370 \\ \hline

\end{tabular}
\end{table}

\end{document}